\documentclass[]{fairmeta} %

\usepackage{amsmath,amsfonts,bm}

\def\eqref#1{equation~\ref{#1}}

\def\1{\bm{1}}

\DeclareMathAlphabet{\mathsfit}{\encodingdefault}{\sfdefault}{m}{sl}
\SetMathAlphabet{\mathsfit}{bold}{\encodingdefault}{\sfdefault}{bx}{n}

\usepackage[utf8]{inputenc}
\usepackage[T1]{fontenc}
\usepackage{lmodern}
\usepackage{microtype}
\usepackage{xspace}

\usepackage{amsmath}
\usepackage{amssymb}
\usepackage{amsfonts}
\usepackage{dsfont}
\usepackage{nicefrac}

\usepackage[dvipsnames,table]{xcolor}
\usepackage{graphicx}
\usepackage{wrapfig}
\usepackage{placeins}

\usepackage{booktabs}
\usepackage{multirow}
\usepackage{multicol}
\usepackage{tabularx}
\usepackage{makecell}
\usepackage{colortbl}

\usepackage[ruled,vlined]{algorithm2e}
\usepackage{listings}

\usepackage[most]{tcolorbox}
\tcbuselibrary{breakable}
\usepackage{mdframed}

\usepackage{enumitem}
\usepackage{caption}
\usepackage{subcaption}
\usepackage{url}

\usepackage{hyperref}

\usepackage{multicol}
\usepackage{multirow}
\usepackage{pifont}
\usepackage{tabularx}
\newcolumntype{R}{>{\raggedleft\arraybackslash}X}
\definecolor{lightblue}{RGB}{220,235,250}
\definecolor{lightgray}{gray}{0.91}

\usepackage[most,skins,theorems]{tcolorbox}
\usepackage{listings}
\newtcolorbox{promptbox}[1][]{
  enhanced, breakable,
  colback=gray!1,      
  colframe=gray!60,    
  coltitle=black,      
  boxrule=2pt,
  arc=10pt,
  left=6pt, right=6pt, top=6pt, bottom=6pt,
  title={#1}, fonttitle=\bfseries,
  attach boxed title to top left={yshift*=-3mm},
  boxed title style={colback=gray!10}
}

\tcbset{
  aibox/.style={
    width=\linewidth,
    top=8pt,
    bottom=4pt,
    colback=inftythink-red!15,
    colframe=inftythink-red,
    colbacktitle=inftythink-red!90!black,
    enhanced,
    center,
    attach boxed title to top left={yshift=-0.1in,xshift=0.15in},
    boxed title style={boxrule=0pt,colframe=white,},
  }
}
\newtcolorbox{AIbox}[2][]{aibox,title=#2,#1}
\newtcolorbox{casebox}[1][]{
  enhanced,
  breakable,                %
  skin=standard,            
  colframe=black!75,        %
  colback=black!2,          %
  colbacktitle=black!75,    %
  coltitle=white,           %
  fonttitle=\bfseries\small,%
  fontupper=\small,         %
  sharp corners,            %
  boxrule=0.8pt,            %
  toptitle=1.5mm, bottomtitle=1.5mm, %
  left=2mm, right=2mm,      %
  top=2mm, bottom=2mm,      %
  title={#1},               %
  colframe=black!75,        %
}

\usepackage{amsthm}
\theoremstyle{plain}

\theoremstyle{definition}

\theoremstyle{remark}

\title{Pause or Fabricate? Training Language Models for Grounded Reasoning}

\author[1]{Yiwen Qiu}
\author[1]{Linjuan Wu}
\author[1]{Yizhou Liu}
\author[1]{Yuchen Yan}
\author[2]{Jin Ma}
\author[2]{Xu Tan}
\author[3]{Yao Hu}
\author[3]{Daoxin Zhang}
\author[1]{Wenqi Zhang}
\author[1]{Weiming Lu}
\author[1]{Jun Xiao}
\author[1\dagger]{Yongliang Shen}

\affiliation[1]{Zhejiang University}
\affiliation[2]{Tencent}
\affiliation[3]{Xiaohongshu Inc.}

\abstract{
Large language models have achieved remarkable progress on complex reasoning tasks. However, they often implicitly fabricate information when inputs are incomplete, producing confident but unreliable conclusions—a failure mode we term \textit{\textbf{ungrounded reasoning}}. We argue that this issue arises not from insufficient reasoning capability, but from the lack of \textit{\textbf{inferential boundary awareness}}—the ability to recognize when the necessary premises for valid inference are missing.
To address this issue, we propose \textbf{\underline{G}rounded \textbf{\underline{R}}easoning via \textbf{\underline{I}}nteractive Reinforcement \textbf{\underline{L}}earning (\textbf{GRIL})}, a multi-turn reinforcement learning framework for grounded reasoning under incomplete information. GRIL decomposes the reasoning process into two stages: clarify and pause, which identifies whether the available information is sufficient, and grounded reasoning, which performs task solving once the necessary premises are established. We design stage-specific rewards to penalize hallucinations, enabling models to detect gaps, stop proactively, and resume reasoning after clarification.
Experiments on GSM8K-Insufficient and MetaMATH-Insufficient show that GRIL significantly improves premise detection (up to 45\%), leading to a 30\% increase in task success while reducing average response length by over 20\%. Additional analyses confirm robustness to noisy user responses and generalization to out-of-distribution tasks.
}

\date{\today}
\metadata[Code]{\url{https://github.com/ZJU-REAL/GRIL}}
\correspondence{\email{\{qiuyiwen,syl\}@zju.edu.cn}}

\begin{document}

\maketitle

\section{Introduction}
\label{sec:intro}

Large language models \cite{deepseekai2025deepseekr1incentivizingreasoningcapability,DBLP:journals/corr/abs-2303-08774,DBLP:conf/icml/ZhouZPLK24,DBLP:journals/corr/abs-2412-15115} have demonstrated remarkable capabilities in complex reasoning tasks. Reinforcement learning post-training paradigms, including RLHF \cite{ouyang2022training} and RLVR \cite{deepseekai2025deepseekr1incentivizingreasoningcapability}, have further improved performance on standard benchmarks where models receive well-formed, information-complete inputs \cite{DBLP:journals/corr/abs-2402-03300,DBLP:journals/corr/abs-2508-15260,yan2025mathfimer}.

However, this evaluation paradigm rests on a strong implicit assumption: all necessary information is provided in a single turn. In real-world interactions, this assumption rarely holds. Users often provide information incrementally, expressing requirements ambiguously, or omitting critical details altogether~\cite{luo2025clarifymtbenchbenchmarkingimprovingmultiturn,gan2024clarq}. In such settings, the challenge is not merely to reason correctly, but to determine \textit{whether reasoning is possible at all} given the available information.

\begin{wrapfigure}{r}{0.5\textwidth}
    \centering
    \includegraphics[width=\linewidth]{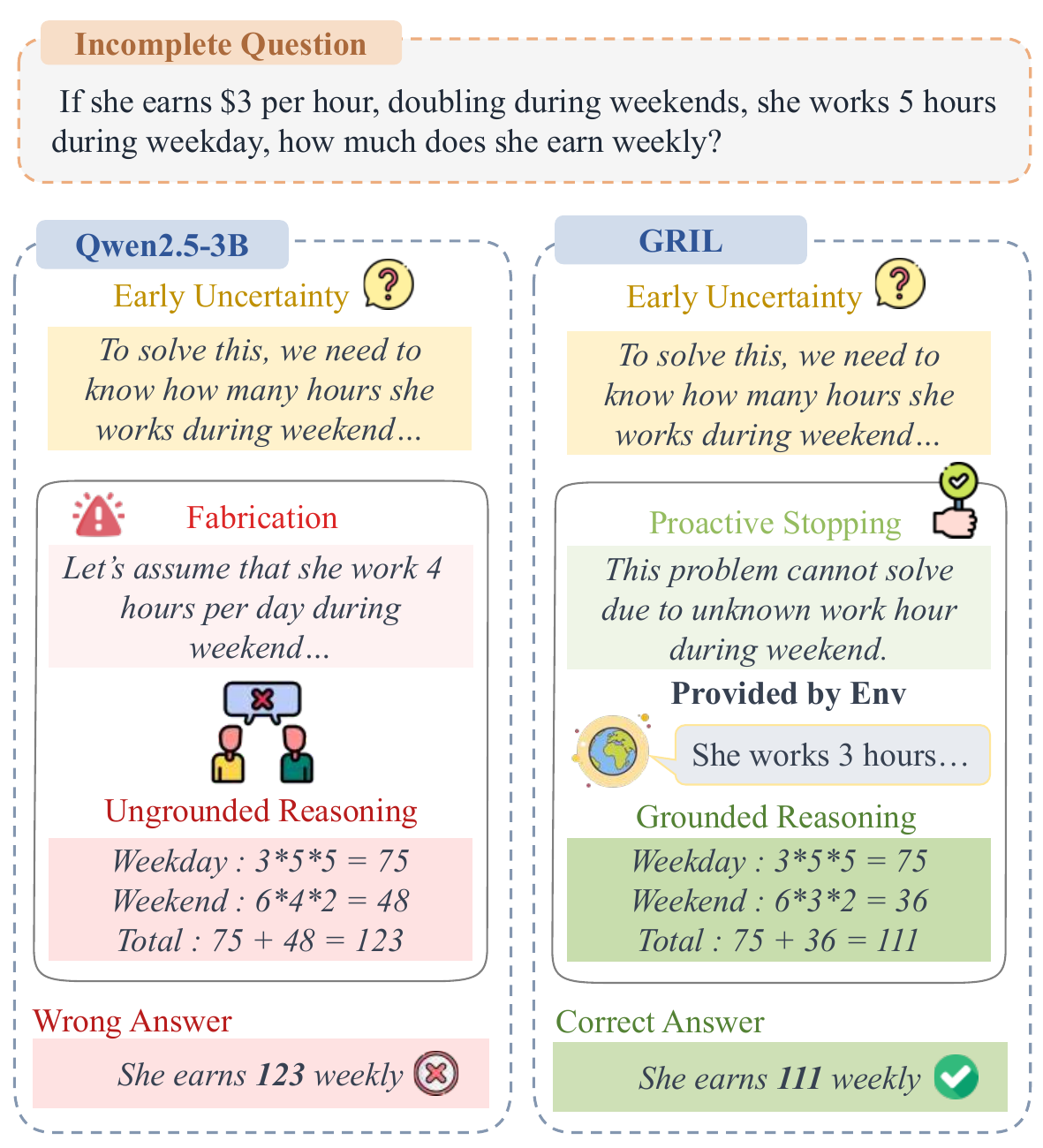}
    \caption{Comparison of reasoning behavior on problems with missing premises. Given an incomplete problem, both models initially detect the information gap. The base model fabricates the missing value and proceeds with ungrounded reasoning, producing an incorrect answer. GRIL-trained model proactively stops and requests clarification, then performs grounded reasoning after receiving the missing premise.
}
    \label{fig:motivation}
\end{wrapfigure}

When confronted with problems that lack necessary premises, current models rather than halting or requesting clarification, they frequently fabricate the missing information and proceed with elaborate reasoning chains built on invented foundations~\cite{DBLP:journals/corr/abs-2504-06514,DBLP:journals/tois/HuangYMZFWCPFQL25}. As illustrated in Figure~\ref{fig:motivation}, we refer to this phenomenon as \textit{ungrounded reasoning}: the continuation of inference without sufficient grounding information. Importantly, ungrounded reasoning is not a failure of logical capability. The failure occurs because the model reasons without checking whether the premises are valid, resulting in outputs that appear coherent yet are fundamentally unreliable~\cite{sui2025stop,wang2025comprehensive}.

Our empirical analysis reveals that ungrounded reasoning follows a characteristic trajectory, as examples shown in Figure~\ref{fig:motivation} (Left). Models often express early uncertainty through phrases such as ``we need to know'' or ``this requires,'' typically within the first half of their responses. Instead of stopping, however, they rapidly transition to fabrication using phrases like ``let us assume'' and then continue generating long chains of baseless 
inference. We capture this behavior using two metrics: \textit{GapRatio}, which measures the proportion of tokens generated after the first uncertainty signal, and \textit{Premise Detection Rate}, which measures how often models correctly identify inputs as incomplete. In preliminary studies, we observe GapRatios exceeding 47.7\% and Premise Detection Rates below 41\%. This suggests that models often override early uncertainty signals, proceeding with extended inference instead of stopping to request missing information.

Why does this behavior persist? Standard reinforcement learning frameworks reward answer production but provide no mechanism to recognize when no valid answer exists. As a result, models learn to answer rather than abstain, especially when trained on well-formed, single-turn problems~\cite{sharma2023towards,rita2024countering}. This implies that ungrounded reasoning reflects a failure of \textit{inferential control}: models cannot reliably decide when to stop and seek clarification~\cite{li2023inference,kadavath2022language,xiong2024can}. We thus cast the problem as a choice between two inferential actions: \textit{Grounded Reasoning}, treating the input as sufficient and answering, or \textit{Clarify and Pause}, identifying missing premises and halting until clarification is obtained~\cite{kuhn2022clam,wang2023interactive}.
Building on this insight, we introduce \textbf{Grounded Reasoning via Interactive Reinforcement Learning (GRIL)}, a multi-turn RL framework that explicitly trains models to recognize the boundaries of valid inference. GRIL decomposes reasoning into \textit{clarify and pause}, and \textit{grounded reasoning}, with an interactive training environment where correct identification of missing information triggers provision of the missing premise. Specifically, in clarify and pause stage, the model interacts with the environment over multiple turns to assess information sufficiency, and is rewarded for proactively stopping and requesting clarification when premises are missing, with a time-decay detection reward to encourage early identification. When model successfully detects information insufficiency, missed premises will be provided by the environment, the model transitions to grounded reasoning stage, where it performs grounded task solving under the completed context and receives a solving reward only upon producing a correct answer. Through stage-specific rewards with temporal decay, GRIL penalizes delayed detection, discourages hallucinated reasoning, and encourages efficient information use.

We evaluate GRIL across multiple model scales on mathematical reasoning benchmarks with systematically removed premises. On GSM8K-Insufficient, GRIL yields substantial improvements on Qwen2.5-1.5B, increasing the Premise Detection Rate from 4.6\% to 90.8\% and the Success Rate from 1.8\% to 61.6\%. At the same time, response length decreases by over 40\%, reflecting earlier termination of ungrounded inference. Notably, GRIL does not degrade performance on complete problems, and achieves gains of up to 7\% on standard GSM8K. Further analyses confirm robustness to noisy user responses and generalization beyond mathematical reasoning.

Our contributions are: (1) We identify \textit{ungrounded reasoning} as a distinct failure mode with quantitative metrics for measurement. (2) We reframe reasoning under uncertainty as sequential decision-making, establishing that inferential boundary awareness is distinct from reasoning capability. (3) We propose GRIL, a multi-turn RL framework with stage-specific rewards for premise detection and grounded solving. (4) Through extensive experiments, we show that GRIL substantially reduces ungrounded reasoning while also improving performance on standard benchmarks.

\section{Related Work}

\paragraph{Reasoning in Large Language Models.} Large language models have demonstrated remarkable reasoning capabilities through various prompting and training strategies \cite{lightman2023let,wang2024math,shinn2023reflexion}. Chain-of-thought prompting \cite{DBLP:conf/nips/Wei0SBIXCLZ22} and its variants \cite{DBLP:conf/nips/YaoYZS00N23,DBLP:conf/aaai/BestaBKGPGGLNNH24,DBLP:conf/iclr/0002WSLCNCZ23} elicit step-by-step reasoning by generating intermediate steps before final answers. Reinforcement learning has further improved reasoning through RLHF \cite{ouyang2022training,DBLP:journals/corr/abs-2204-05862} and process-based reward models \cite{DBLP:journals/corr/abs-2406-06592,DBLP:journals/corr/abs-2503-21295,khalifa2025process}. However, recent studies reveal limitations when inputs are incomplete. \citet{DBLP:journals/corr/abs-2504-06514,liu2025reversethinkingenhancesmissing} show that reasoning models tend to exhibit overthinking when facing problems with missing premises, often fabricating assumptions rather than recognizing information insufficiency. This tendency appears more pronounced in models trained with RL, where optimization pressure to produce answers may override uncertainty recognition. Our work addresses this by explicitly training models to detect missing premises and request clarification.

\paragraph{Multi-Turn RL for Interaction and Reasoning.} Multi-turn interaction \cite{li2025search,jin2025search} presents challenges for language models. \citet{DBLP:journals/corr/abs-2505-06120} demonstrate that when information is distributed across multiple turns, model performance degrades significantly compared to single-turn settings, with drops averaging 30\%. Missing premise scenarios represent a special case where critical information arrives only after explicit request. Multi-turn RL has been explored for dialogue \cite{DBLP:conf/nips/MadaanTGHGW0DPY23,gao2025modeling,jiang2025pagmultiturnreinforcedllm} and tool-augmented agents \cite{DBLP:conf/nips/SchickDDRLHZCS23,DBLP:conf/iclr/YaoZYDSN023,DBLP:journals/corr/abs-2503-01940}, with frameworks like RAGEN \cite{DBLP:journals/corr/abs-2504-20073} addressing long-horizon optimization. Clarification behavior has been studied in conversational AI \cite{DBLP:conf/coling/HuWWLM20,zhang2024modeling,hao2025rl}, while \citet{liu2025reversethinkingenhancesmissing} propose reverse reasoning to detect missing information. Our work differs by framing the problem as learning inferential control through interactive RL, training models to detect, stop, and integrate missing information within a unified framework.

\begin{figure*}[t]
    \centering
    \includegraphics[width=\textwidth]{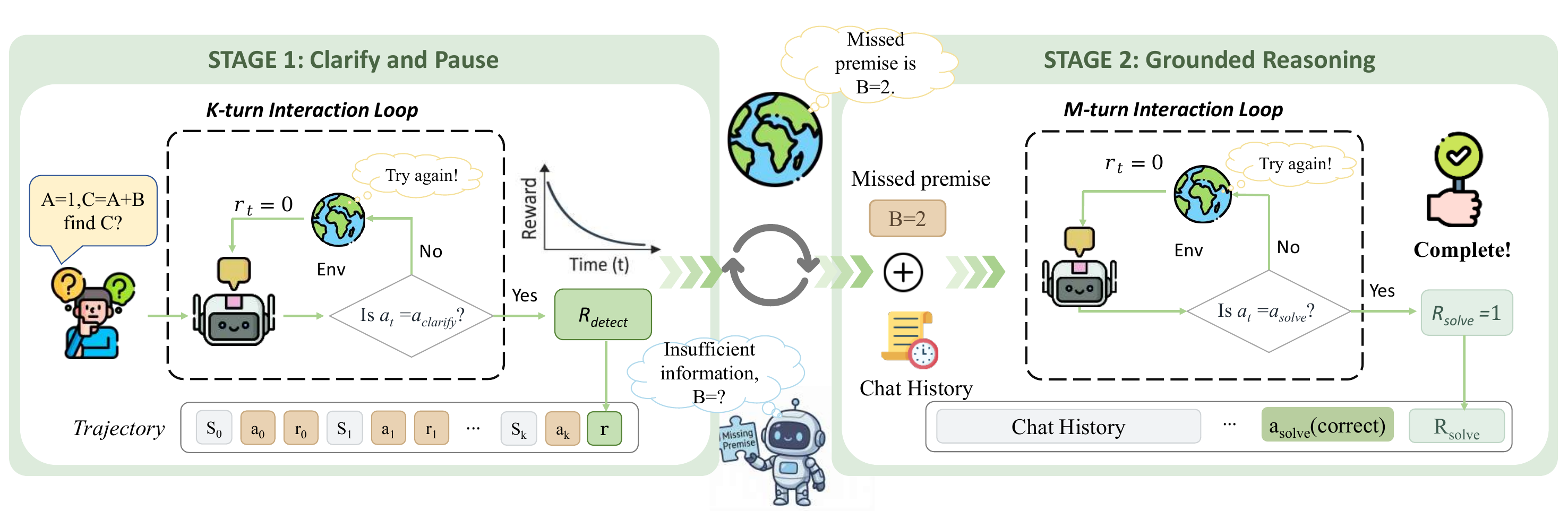}
    \caption{\textbf{Overview of the GRIL framework.} The training process consists of two stages. \textbf{Stage 1 (Clarify and Pause)}: Given an incomplete problem, the model engages in a multi-turn interaction loop. If the model attempts to solve without sufficient information, it receives negative feedback and zero reward. When the model correctly identifies insufficient information and requests clarification, it receives a detection reward $R_{\text{detect}} = \gamma^t$ that decays with the number of interaction turns, encouraging early detection. \textbf{Stage 2 (Grounded Reasoning)}: The environment provides the missing premise, which is concatenated with the chat history. The model integrates this information to produce the final answer. A correct solution yields the solving reward $R_{\text{solve}}$, otherwise the model may retry until maximum turns are reached.}
    \label{fig:method}
\end{figure*}

\section{Method}

We present \textbf{Grounded Reasoning via Interactive Reinforcement Learning (GRIL)}, a multi-turn RL framework that trains language models to recognize inferential boundaries and reason only when sufficient information is available. As illustrated in Figure~\ref{fig:method}, the core insight is to decompose reasoning into two distinct stages: \textit{clarify and pause}, where the model determines whether the input contains sufficient information, and \textit{grounded reasoning}, where the model produces a solution only when necessary premises are present. We formalize this as an interactive decision process with stage-specific rewards that penalize ungrounded inference and encourage appropriate clarification-seeking behavior.

\subsection{Problem Formulation}

Consider a reasoning task where the model receives a problem statement that may or may not contain all information necessary for solution. We formalize this as a multi-turn Markov Decision Process $\mathcal{M} = \langle \mathcal{S}, \mathcal{A}, \mathcal{P}, \mathcal{R}, \gamma \rangle$.

\paragraph{State Space.} At turn $t$, the state $s_t$ comprises the complete interaction history:
\begin{equation}
    s_t = (u_1, a_1, u_2, a_2, \ldots, u_t)
\end{equation}
where $u_i$ denotes environment messages (initial problem, feedback, or clarifying information) and $a_i$ denotes model responses.

\paragraph{Action Space.} Although model outputs are natural language sequences, they serve functionally distinct roles in our framework. We abstract model behavior into two inferential actions:
\begin{itemize}[leftmargin=2em]
    \item \textbf{Solve} ($a_{\text{solve}}$): The model commits to the current information as sufficient and produces a solution attempt.
    \item \textbf{Clarify} ($a_{\text{clarify}}$): The model identifies that critical information is missing and explicitly requests the needed premise.
\end{itemize}

This abstraction captures the key insight that \textit{choosing to reason} is itself a decision contingent on available information. Ungrounded reasoning occurs precisely when models select $a_{\text{solve}}$ in states where $a_{\text{clarify}}$ is appropriate.

\paragraph{Problem Types.} We denote the set of incomplete problems (with missing premises) as $\mathcal{D}_{\text{inc}}$ and complete problems (with sufficient information) as $\mathcal{D}_{\text{comp}}$. For $q \in \mathcal{D}_{\text{inc}}$, there exists a missing premise $p_q$ that, when provided, renders the problem solvable. The goal is to learn a policy $\pi_\theta$ that selects appropriate actions based on input completeness.

\subsection{Interactive Environment}

A central challenge in training inferential control is that action quality cannot be assessed from immediate outputs alone. A model that selects $a_{\text{solve}}$ on an incomplete problem may generate internally coherent reasoning, yet the entire inference chain lacks validity. To enable learning from the consequences of inferential choices, we design an interactive environment with asymmetric feedback dynamics.

\paragraph{Transition Dynamics.} The environment responds differently based on problem type and model action. For incomplete problems $q \in \mathcal{D}_{\text{inc}}$, selecting $a_{\text{clarify}}$ with correct detection triggers provision of the missing premise $p_q$, while selecting $a_{\text{solve}}$ yields negative feedback. For complete problems $q \in \mathcal{D}_{\text{comp}}$, selecting $a_{\text{solve}}$ with a correct answer terminates the episode successfully, while selecting $a_{\text{clarify}}$ receives indication that no additional information is needed.

Formally, let $\mathcal{T}$ denote the transition function and $\oplus$ denotes concatenation to the dialogue history. For $q \in \mathcal{D}_{\text{inc}}$:
\begin{equation}
    \mathcal{T}(s_t, a_t) = 
    \begin{cases}
        s_t \oplus p_q & a_t = a_{\text{clarify}} \\
        s_t \oplus \texttt{[neg.]} & a_t = a_{\text{solve}}
    \end{cases}
\end{equation}
For $q \in \mathcal{D}_{\text{comp}}$:
\begin{equation}
    \mathcal{T}(s_t, a_t) =
    \begin{cases}
        \texttt{[done]} & \text{if } a_t = a_{\text{solve}} \\
        s_t \oplus \texttt{[unc.]} & \text{if } a_t = a_{\text{clarify}}
    \end{cases}
\end{equation}
where \texttt{[neg.]} denotes negative feedback, \texttt{[done]} denotes successful termination, and \texttt{[unc.]} denotes unnecessary clarification. This design creates a learning signal where correct premise detection leads to receiving the missing information, enabling the model to complete valid reasoning. The asymmetry ensures that clarification is beneficial only when genuinely warranted, preventing degeneration into overly conservative strategies.

\subsection{Stage-Specific Reward Design}

We design a reward function that explicitly decomposes the reasoning process into \textit{clarify and pause}, and \textit{grounded reasoning}, with separate incentives for each stage.

\paragraph{Stage 1: Pause and Clarify.} For incomplete problems, we reward early and accurate identification of missing information. Let $n_{\text{prior}}$ denote the number of interaction turns generated before the model requests clarification and stops proactively. The detection reward is:
\begin{equation}
    R_{\text{detect}} = r_{\text{base}} \cdot \gamma_d^{n_{\text{prior}}}
\end{equation}
where $r_{\text{base}}$ is the base detection reward and $\gamma_d \in (0, 1)$ is a temporal decay factor. This formulation directly addresses the "early suspicion, late action" pattern: models that recognize missing information early receive higher rewards than those that generate extensive ungrounded tokens before requesting clarification. The exponential decay provides strong incentive to minimize $n_{\text{prior}}$.

\paragraph{Stage 2: Grounded Reasoning.} After receiving clarifying information, we reward successful problem completion:
\begin{equation}
    R_{\text{solve}} = r_{\text{correct}} \cdot \mathds{1}[\text{answer is correct}]
\end{equation}
This ensures that premise detection is not an end in itself but a means toward successful task completion. Models must learn to integrate newly provided information with the original context to produce valid solutions.

\paragraph{Complete Problem Handling.} For problems in $\mathcal{D}_{\text{comp}}$, we apply standard outcome-based rewards with a penalty for unnecessary clarification:
\begin{equation}
    R_{\text{comp}} = r_{\text{correct}} \cdot \mathds{1}[\text{correct}] - \lambda \cdot \mathds{1}[\text{unc.}]
\end{equation}
where $\mathds{1}[\text{unc.}]$ indicates unnecessary clarification on complete problems. The penalty coefficient $\lambda$ prevents models from adopting a trivial strategy of always requesting clarification regardless of input completeness.

\paragraph{Overall Objective.} The total reward for a trajectory $\tau$ on problem $q$ is:
\begin{equation}
    R(q, \tau) = 
    \begin{cases}
        \alpha \cdot R_{\text{detect}} + \beta \cdot R_{\text{solve}} & \text{if } q \in \mathcal{D}_{\text{inc}} \\
        R_{\text{comp}} & \text{if } q \in \mathcal{D}_{\text{comp}}
    \end{cases}
\end{equation}
where $\alpha$ and $\beta$ control the relative importance of detection versus solving. We optimize using Proximal Policy Optimization (PPO) \cite{DBLP:journals/corr/SchulmanWDRK17} with KL regularization:
\begin{equation}
    \mathcal{J}(\theta) = \mathds{E}_{q \sim \mathcal{D}, \tau \sim \pi_\theta} \left[ R(q, \tau) \right] - \beta_{\text{KL}} \cdot D_{\text{KL}}(\pi_\theta \| \pi_{\text{ref}})
\end{equation}
where $\pi_{\text{ref}}$ is the reference model and $\beta_{\text{KL}}$ controls deviation from the initial policy.

\subsection{Training Protocols}

\paragraph{Data Construction.} We construct incomplete problems by systematically removing critical premises from well-formed mathematical problems. For each problem, we identify sentences containing numerical values, remove one such sentence, and verify via automated checking that the modified problem becomes unsolvable without the removed information. The removed sentence is retained as $p_q$ for use during interactive training. We also randomly sampled a subset of questions for manual annotation verification. Details are provided in Appendix.

\paragraph{Data Composition.} Training data comprises a balanced mixture of incomplete and complete problems with ratio 1:1. This balance is critical: training exclusively on incomplete problems would bias models toward over-detection, while training only on complete problems provides no signal for learning clarification behavior.

\paragraph{Output Format.} We enforce structured generation with \texttt{<think>...</think>} tags for reasoning traces and \texttt{<answer>...</answer>} tags for final responses. When requesting clarification, models output "insufficient information" in the answer field . This structured format enables unambiguous identification of inferential actions.

\newcolumntype{L}{>{\raggedright\arraybackslash}X}
\newcolumntype{C}{>{\centering\arraybackslash}X}
\definecolor{graybg}{gray}{0.95}
\section{Experiments}

\subsection{Experimental Setup}

\paragraph{Models and Training.} We evaluate across four model scales: Qwen2.5-1.5B-Instruct, Qwen2.5-3B-Instruct \cite{DBLP:journals/corr/abs-2412-15115}, Qwen3-0.6B, and Qwen3-1.7B \cite{DBLP:journals/corr/abs-2505-09388}. All models are trained using the verl \cite{DBLP:conf/eurosys/ShengZYWZZPL025} framework with maximum turns $T=4$, reward weights $\alpha=0.3$, $\beta=0.7$, temporal decay $\gamma_d=0.5$, KL coefficient $\beta_{\text{KL}}=0.01$, and penalty coefficient$\lambda=2$. Training data consists of incomplete and complete problems mixed at 1:1 ratio.

\paragraph{Datasets.} We construct two evaluation benchmarks: \textbf{GSM8K-Insufficient} and \textbf{MetaMATH-Insufficient}, derived from GSM8K \cite{DBLP:journals/corr/abs-2110-14168} and MetaMATH \cite{DBLP:conf/iclr/YuJSYLZKLWL24} by systematically removing critical premises.  Notably, there is no overlap between the training and test sets. We also evaluate on standard GSM8K and MATH500 to assess reasoning capability on complete problems. Details are in Appendix~\ref{app:data_report_sec5}.

\paragraph{Baselines.} We compare against three settings: (1) \textbf{Base}: the pretrained model with zero-shot prompting; (2) \textbf{Prompt}: the base model with explicit instructions indicating that problems may contain missing information; (3) \textbf{SFT}: supervised fine-tuning on the same data distribution as GRIL. Detailed SFT training process is displayed in Appendix ~\ref{sec:baselines}

\paragraph{Metrics.} We report four metrics: \textit{Success Rate (SR)}, the proportion of problems solved correctly after clarification if needed; \textit{Premise Detection (PD)}, the proportion of incomplete problems correctly identified; \textit{Numbers of Iteraction Turns (NT)}, the average interaction turns required; and \textit{Response Length}, the average tokens generated.

\begin{table*}[h]
\centering
\small
\setlength{\tabcolsep}{3pt}
\begin{tabularx}{\textwidth}{l *{7}{L} C}
\toprule
\multicolumn{1}{c}{\multirow{2}{*}{\textbf{Model}}} & \multicolumn{4}{c}{GSM8K-Insufficient} & \multicolumn{4}{c}{Metamath-Insufficient} \\
\cmidrule(lr){2-5} \cmidrule(lr){6-9}

 & SR $\uparrow$ & PD $\uparrow$ & NT $\downarrow$ & Length $\downarrow$ & SR $\uparrow$ & PD $\uparrow$ & NT $\downarrow$ & Length $\downarrow$ \\
\midrule

\multicolumn{9}{l}{\textbf{Qwen2.5-1.5B-Instruct}} \\
\quad Base Model & 1.8 & 4.6 & 3.828 & 810 & 2.2 & 4.2 & 3.792 & 937 \\
\quad w/ Prompt  & 21.0 & 45.7 & 3.303 & 579 & 19.3 & 48.1 & 3.426 & 743 \\
\quad w/ SFT     & 31.6 & 64.6 & 3.202 & 750 & 31.1 & 60.4 & 3.248 & 999 \\
\rowcolor{graybg} \quad \textbf{GRIL(Ours)} & \textbf{61.6} & \textbf{90.8} & \textbf{2.913} & \textbf{479} & \textbf{58.4} & \textbf{88.2} & \textbf{2.941} & \textbf{581} \\

\multicolumn{9}{l}{\textbf{Qwen2.5-3B-Instruct}} \\
\quad Base Model & 20.6 & 28.0 & 3.665 & 887 & 15.5 & 20.9 & 3.722 & 1091 \\
\quad w/ Prompt  & 54.9 & 81.5 & 2.919 & 635 & 51.5 & 77.1 & 2.985 & 824 \\
\quad w/ SFT     & 62.7 & \textbf{90.3} & 2.706 & 575 & 56.1 & 83.9 & 2.800 & 750 \\
\rowcolor{graybg} \quad \textbf{GRIL(Ours)} & \textbf{73.5} & 88.0 & \textbf{2.481} & \textbf{448} & \textbf{72.5} & \textbf{86.6} & \textbf{2.473} & \textbf{624} \\

\multicolumn{9}{l}{\textbf{Qwen3-0.6B}} \\
\quad Base Model & 16.0 & 24.2 & 3.737 & 1668 & 14.7 & 25.6 & 3.793 & 1907 \\
\quad w/ Prompt  & 44.6 & 79.2 & \textbf{2.763} & 1112 & 40.6 & 75.7 & \textbf{2.869} & 1345 \\
\quad w/ SFT     & 48.4 & 82.4 & 2.971 & \textbf{832}  & 41.3 & \textbf{81.8} & 3.077 & \textbf{1003} \\
\rowcolor{graybg} \quad \textbf{GRIL(Ours)} & \textbf{52.3} & \textbf{84.6} & 2.850 & 1269 & \textbf{45.2} & 79.2 & 3.025 & 1616 \\

\multicolumn{9}{l}{\textbf{Qwen3-1.7B}} \\
\quad Base Model & 41.3 & 52.6 & 3.271 & 1271 & 62.0 & 90.5 & 2.585 & 929 \\
\quad w/ Prompt  & 66.7 & 88.9 & 2.481 & 744  & 64.2 & 86.1 & 2.609 & 966 \\
\quad w/ SFT     & 63.6 & 92.8 & 2.482 & 704  & 63.0 & 90.3 & 2.531 & 874 \\
\rowcolor{graybg} \quad \textbf{GRIL(Ours)} & \textbf{72.8} & \textbf{96.5} & \textbf{2.348} & \textbf{376}  & \textbf{65.8} & \textbf{95.2} & \textbf{2.488} & \textbf{502} \\
\bottomrule
\end{tabularx}
\caption{Results on GSM8K-Insufficient and MetaMATH-Insufficient across model scales. SR: Success Rate (\%), PD: Premise Detection (\%), NT: Numbers of Interaction Turns, Length: Response Length in tokens.}
\label{tab:main_results}
\end{table*}

\subsection{Main Results}

Table~\ref{tab:main_results} presents results on GSM8K-Insufficient and MetaMATH-Insufficient across model scales. GRIL consistently outperforms all baselines on both benchmarks.

\paragraph{GRIL Dramatically Improves Premise Detection and Task Success.} Across all model scales, GRIL achieves substantial gains in both identifying incomplete inputs and ultimately solving problems. On Qwen2.5-1.5B, Success Rate improves from 1.8\% (Base) to 61.6\%, a 34$\times$ improvement, while Premise Detection rises from 4.6\% to 90.8\%. The gains persist at larger scales: Qwen2.5-3B achieves 73.5\% SR compared to 62.7\% for SFT, and Qwen3-1.7B reaches 72.8\% SR with 96.5\% detection rate. 

\paragraph{GRIL Outperforms Supervised Learning on Identical Data.} GRIL consistently outperforms SFT across various settings. On Qwen2.5-1.5B, the SR gap is 30 percentage points (61.6\% vs 31.6\%); on MetaMATH-Insufficient, GRIL achieves 72.5\% compared to 56.1\% for SFT on Qwen2.5-3B. This shows that multi-turn RL lets the model learn from feedback after it makes a choice, and this kind of learning signal can’t be obtained from imitation learning

\paragraph{Reduced Ungrounded Reasoning Yields Efficiency Gains.} GRIL not only improves accuracy but also reduces inference cost. Response length decreases substantially: Qwen2.5-1.5B drops from 810 to 479 tokens (41\% reduction), Qwen3-1.7B from 1271 to 376 tokens (70\% reduction). The number of interaction turns also decreases, indicating earlier detection of missing premises. One exception is Qwen3-0.6B, where response length increases despite accuracy gains. This suggests a capacity threshold: models below approximately 1B parameters may require additional tokens to articulate clarifications and integrate multi-turn context, while larger models learn to be both accurate and concise.

\section{Analysis}
\label{sec:Detaied Analysis}

\begin{wrapfigure}{r}{0.5\textwidth}
    \centering
    \includegraphics[width=1\linewidth]{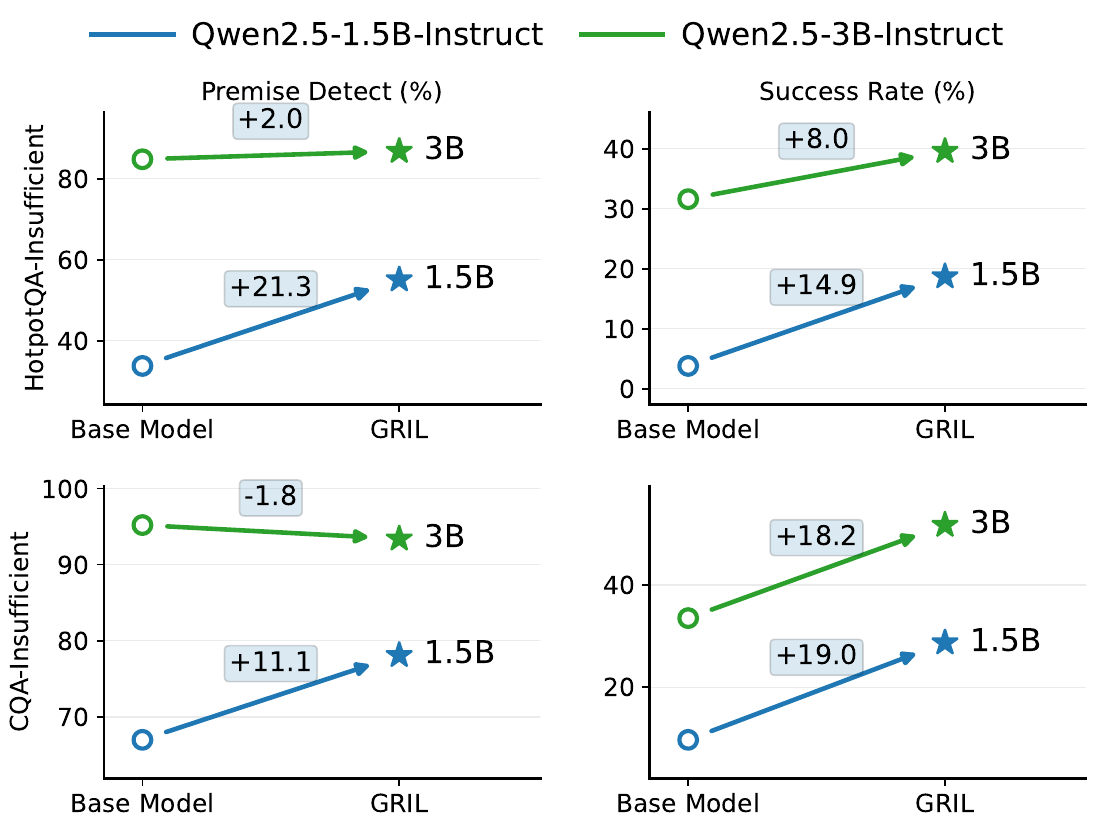} 
    \caption{Out-of-distribution generalization on HotpotQA-Insufficient and CQA-Insufficient. Arrows indicate improvement trajectories from base models to GRIL-trained variants.}
    \label{fig:analysis_five}
\end{wrapfigure}

We conduct detailed analyses to understand the mechanisms behind GRIL's improvements. Our investigation covers five aspects: reduction in ungrounded reasoning, performance on complete problems, discrimination between complete and incomplete inputs, decomposition of performance gains, and generalization capabilities.
\begin{figure}[htbp]
    \centering
    \begin{minipage}[b]{0.48\textwidth}
        \centering
        \includegraphics[width=\linewidth]{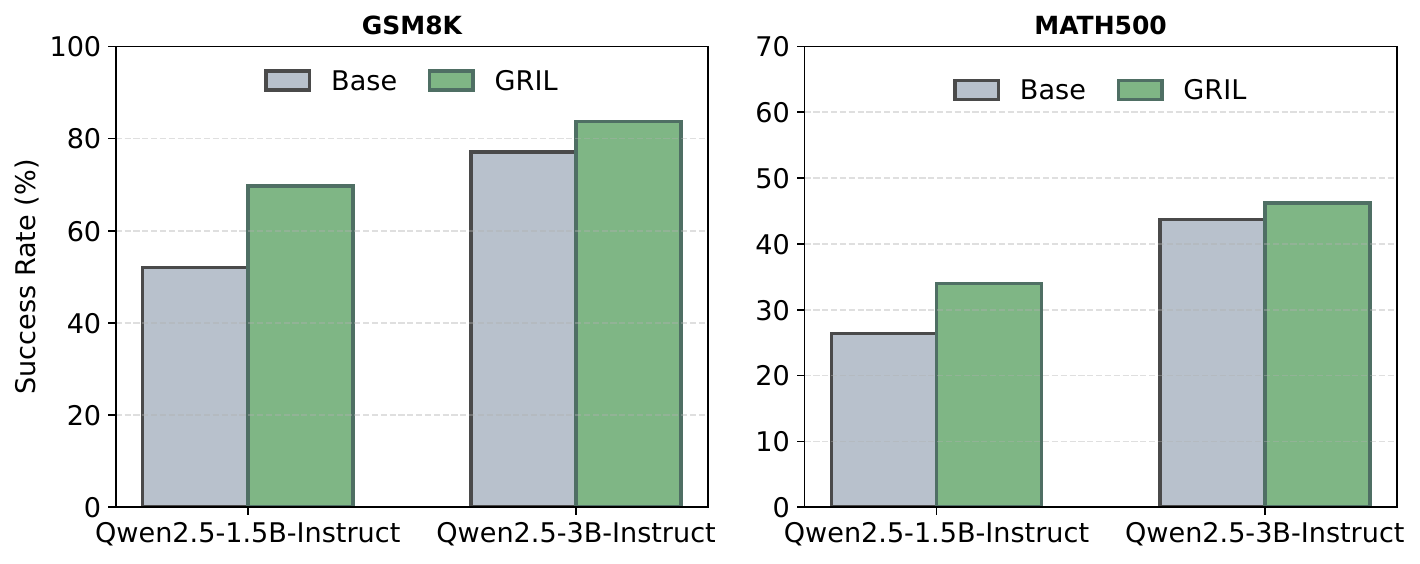}
        \caption{Success Rate on standard complete benchmarks (GSM8K and MATH500) comparing base models with GRIL-trained variants.}
        \label{fig:analysis_two}
    \end{minipage}
    \hfill %
    \begin{minipage}[b]{0.48\textwidth}
        \centering
        \includegraphics[width=\linewidth]{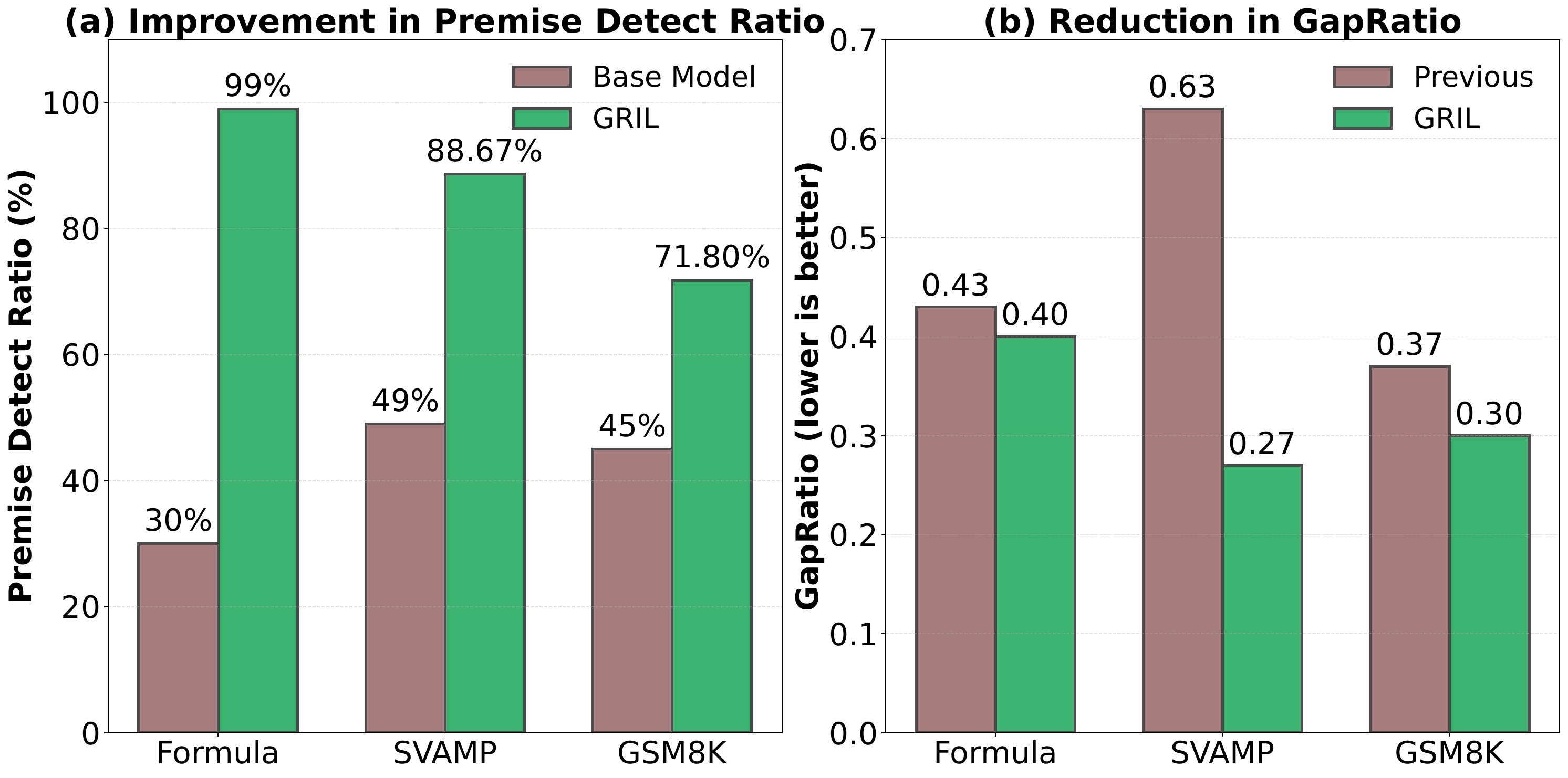}
        \caption{Premise Detection Rate and GapRatio before and after GRIL training on three public benchmarks from \citet{DBLP:journals/corr/abs-2504-06514}.}
        \label{fig:analysis_one}
    \end{minipage}
    
\end{figure}
\paragraph{GRIL Substantially Reduces Ungrounded Reasoning.}
To quantify the severity of ungrounded reasoning, we measure the \textit{GapRatio}\footnote{Formally, $\text{GapRatio} = (T_{\text{total}} - T_{\text{suspect}}) / T_{\text{total}}$, where $T_{\text{total}}$ is the total tokens generated and $T_{\text{suspect}}$ is the token position where the model first expresses uncertainty.}: the proportion of tokens generated after the model first expresses uncertainty. A higher ratio indicates more severe ungrounded inference. We evaluate on three public datasets from \citet{DBLP:journals/corr/abs-2504-06514}. As shown in Figure~\ref{fig:analysis_one}, GRIL reduces this metric across all evaluation datasets. On SVAMP, the GapRatio drops from 0.63 to 0.27; on Formula, from 0.43 to 0.40; on GSM8K, from 0.37 to 0.30. Simultaneously, Premise Detection Rate increases substantially: Formula rises from 30\% to 99\%,SVAMP from 49\% to 88.7\%, and GSM8K from 45\% to 71.8\%. These results confirm that GRIL trains models to recognize inferential boundaries early and act on that recognition, rather than suppressing uncertainty signals in favor of continued generation.
\begin{table}[htbp]
    \centering
    
    \begin{minipage}[t]{0.48\textwidth}
        \centering
        \scriptsize
        \setlength{\tabcolsep}{4pt} 
        \begin{tabular}{lccc}
            \toprule
            Model & Recall & Precision & F1 \\
            \midrule
            \textbf{Qwen2.5-1.5B-Instruct} & 0.419 & 0.951 & 0.581 \\
            \quad + GRIL & \textbf{0.861} & \textbf{0.961} & \textbf{0.908} \\
            \midrule
            \textbf{Qwen2.5-3B-Instruct} & 0.756 & 0.968 & 0.847 \\
            \quad + GRIL & \textbf{0.886} & \textbf{0.981} & \textbf{0.931} \\
            \bottomrule
        \end{tabular}
        \caption{Premise detection performance (Recall, Precision, F1) on a mixed dataset containing both complete and incomplete queries.}
        \label{tab:premise_detection_results}
    \end{minipage}
    \hfill %
    \begin{minipage}[t]{0.48\textwidth}
        \centering
        \resizebox{\linewidth}{!}{
            \begin{tabular}{lcccc}
                \toprule
                \multirow{2}{*}{Model} & \multicolumn{2}{c}{GSM8K-Insufficient} & \multicolumn{2}{c}{MetaMath-Insufficient} \\
                \cmidrule(lr){2-3} \cmidrule(lr){4-5} 
                & DCR (\%) & NCR (\%) & DCR (\%) & NCR (\%) \\
                \midrule
                \textbf{Qwen2.5-1.5B-Instruct} & 37.3 & 36.9 & 36.0 & 37.3 \\
                \quad + GRIL & \textbf{66.9} & \textbf{48.2} & \textbf{65.7} & \textbf{57.9} \\
                \midrule
                \textbf{Qwen2.5-3B-Instruct} & 63.7 & 54.2 & 67.3 & 52.8 \\
                \quad + GRIL & \textbf{80.8} & \textbf{63.6} & \textbf{83.7} & \textbf{57.4} \\
                \bottomrule
            \end{tabular}
        }
        \caption{Decomposition of Success Rate into Detected Correct Ratio (DCR) and Non-detected Correct Ratio (NCR) on GSM8K-Insufficient and MetaMATH-Insufficient.}
        \label{tab:dcr_ncr_results}
    \end{minipage}
\end{table}
\paragraph{GRIL Improves Performance on Complete Problems.}
A natural concern is whether training for premise detection compromises reasoning ability on well-formed inputs. Figure~\ref{fig:analysis_two} demonstrates that GRIL not only preserves but enhances standard reasoning performance.On GSM8K, Qwen2.5-1.5B improves from 52.0\% to 69.7\% (+17.7 points), while Qwen2.5-3B rises from 77.1\% to 83.7\%. On the more challenging MATH500 benchmark, gains are also consistent: 26.4\% to 34.0\% for the 1.5B model, 43.7\% to 46.2\% for the 3B model. We hypothesize that learning to distinguish sufficient from insufficient information induces more careful reasoning: models that must decide whether to proceed learn to assess problem structure more thoroughly, benefiting downstream solving even when all information is present.
\paragraph{GRIL Robustly Distinguishes Complete and Incomplete Inputs.}
Effective inferential control requires high sensitivity to missing premises (Recall) without false alarms on complete problems (Precision). We construct a mixed evaluation set containing equal proportions of complete and incomplete queries and measure standard classification metrics.
\newcolumntype{Y}{>{\centering\arraybackslash}X}
\begin{wraptable}{r}{0.5\textwidth}
    \centering
    \scriptsize
    \setlength{\tabcolsep}{3pt} 
    \vspace{0.5em}
    \begin{tabularx}{\linewidth}{lYY}
        \toprule
        Model & Noisy Feedback & Uninformative Response \\
        \midrule
        \textbf{Qwen2.5-1.5B-Instruct} & 12.8 & 62.5 \\
        \quad + GRIL & \textbf{47.2} & \textbf{86.7} \\
        \midrule
        \textbf{Qwen2.5-3B-Instruct} & 51.4 & 89.1 \\
        \quad + GRIL & \textbf{69.7} & \textbf{89.5} \\
        \bottomrule
    \end{tabularx}
    \caption{Success Rate (\%) under noisy feedback and uninformative user response conditions.}
    \label{tab:robustness}
\end{wraptable}
As shown in Table~\ref{tab:premise_detection_results}, GRIL achieves substantial Recall gains: Qwen2.5-1.5B improves from 0.419 to 0.861 (+105\%), Qwen2.5-3B from 0.756 to 0.886. Crucially, Precision remains above 0.96 and even improves slightly, indicating that GRIL does not achieve sensitivity by becoming overly conservative. The resulting F1 scores (0.908 and 0.931) demonstrate that GRIL learns a precise decision boundary, effectively filtering incomplete queries while maintaining grounded reasoning for solvable problems.
\paragraph{Both Detection and Integration Capabilities Improve.}
To decompose performance gains, we design an experiment where models first respond normally, then receive forced environmental feedback regardless of their initial action. We analyze two conditional success rates: \textit{Detected Correct Ratio (DCR)}\footnote{ DCR=$P(\text{success} \mid \text{detected})$}, measuring success after correct premise identification, and \textit{Non-detected Correct Ratio (NCR)}\footnote{ NCR=$P(\text{success} \mid \text{not detected})$}, measuring recovery via forced feedback despite initial detection failure. As shown in Table~\ref{tab:dcr_ncr_results}, GRIL improves both metrics. DCR for Qwen2.5-1.5B nearly doubles (37.3\% $\to$ 66.9\%), validating the effectiveness of the clarify-then-solve cycle. 
NCR also improves substantially (36.9\% $\to$ 48.2\%), indicating that GRIL enhances general context integration capabilities beyond just detection. This dual improvement explains why GRIL outperforms baselines by large margins: it strengthens both the ability to pause appropriately and the ability to reason effectively once sufficient information is available.
\paragraph{GRIL Generalizes to Out-of-Distribution Domains.}
To assess whether learned inferential control transfers beyond mathematical reasoning, we evaluate on modified HotpotQA (multi-hop reasoning) and CommonsenseQA (commonsense reasoning) datasets~\citep{DBLP:conf/emnlp/Yang0ZBCSM18, DBLP:journals/tacl/AroraLFKR23}, 
where we apply the same premise removal procedure used for mathematical benchmarks. As shown in Figure~\ref{fig:analysis_five}, GRIL consistently improves both Premise Detection and Success Rate in these unseen domains. Smaller models (1.5B) show particularly pronounced gains, 
with Success Rate
improvements exceeding 18 percentage points on CQA. The improvement trajectories reveal that GRIL pushes models toward the upper-right region of the detection-success space across both in-domain and out-of-domain tasks. This transfer suggests that GRIL instills a general capacity for inferential boundary awareness rather than domain-specific heuristics.
\paragraph{GRIL Is Robust to Noisy User Interactions.}
Real-world interactions are imperfect: users may provide irrelevant information or fail to supply requested premises. We simulate two challenging conditions: \textit{Noisy Feedback}, where clarifications are mixed with irrelevant daily conversation or off-topic remarks, and \textit{Uninformative Response}, where the user replies with evasive answers such as ``I don't know'' or ``Don't ask me.'' As shown in Table~\ref{tab:robustness}, GRIL substantially improves resilience in both settings. Under noisy feedback, Qwen2.5-1.5B improves from 12.8\% to 47.2\%, indicating learned ability to filter distractions and extract relevant information. Under uninformative responses, performance rises from 62.5\% to 86.7\%, demonstrating graceful termination instead of hallucination when clarification is unavailable. These results confirm that GRIL's benefits extend beyond idealized settings to realistic, noisy human-AI collaboration scenarios.

\section{Conclusion}
In this paper, we identify ungrounded reasoning as a critical failure mode where language models fabricate missing premises and proceed with structurally invalid inference rather than recognizing inferential boundaries. We propose GRIL, a multi-turn reinforcement learning framework that explicitly trains models to distinguish between situations requiring reasoning versus clarification. Through stage-specific rewards that penalize delayed detection and encourage effective information integration, GRIL teaches models when to pause and when to proceed. Extensive experiments demonstrate substantial improvements in both premise detection and task success, with gains generalizing to out-of-distribution domains and noisy interactions. 
\section{Limitations}
Our results are obtained in an idealized interactive setting where an oracle-like environment can judge insufficiency and provide the correct missing premise after a clarification action; real users may refuse, be uncertain, or provide noisy/partial information, so gains may not fully transfer to real-scenario dialogues. In addition, our insufficient queries are synthetically constructed by deleting constraints from math datasets, which may introduce artifacts and cover only a narrow set of missing-information types, limiting generalization beyond arithmetic-style problems. Finally, our implementation relies on a structured output format to parse actions, and we evaluate efficiency mostly with turn/length proxies rather than directly measuring clarification quality and real user cost.Future work should move beyond one-shot, information-complete problem solving to realistic multi-turn conversations where users reveal constraints gradually, requiring models to proactively elicit missing details while handling refusals, uncertainty, and inconsistency. For many open-ended, preference-sensitive tasks, “missing premises” are subjective goals and trade-offs rather than a single correct constraint. This calls for evaluations that directly measure clarification utility and user burden, and training setups robust to noisy, ambiguous, preference-driven feedback.

\section*{Acknowledgment}

This work was supported by National Natural Science Foundation of China (No. 62506332), National Natural Science Foundation of China (No. 62436007), CCF-Tencent Rhino-Bird Open Research Fund, and ZJU Kunpeng\&Ascend Center of Excellence.
\bibliography{main,GRIL}

\begin{thebibliography}{54}
\providecommand{\natexlab}[1]{#1}
\providecommand{\url}[1]{\texttt{#1}}
\expandafter\ifx\csname urlstyle\endcsname\relax
  \providecommand{\doi}[1]{doi: #1}\else
  \providecommand{\doi}{doi: \begingroup \urlstyle{rm}\Url}\fi

\bibitem[Arora et~al.(2023)Arora, Lewis, Fan, Kahn, and R{\'{e}}]{DBLP:journals/tacl/AroraLFKR23}
Simran Arora, Patrick S.~H. Lewis, Angela Fan, Jacob Kahn, and Christopher R{\'{e}}.
\newblock Reasoning over public and private data in retrieval-based systems.
\newblock \emph{Trans. Assoc. Comput. Linguistics}, 11:\penalty0 902--921, 2023.
\newblock \doi{10.1162/TACL\_A\_00580}.
\newblock URL \url{https://doi.org/10.1162/tacl\_a\_00580}.

\bibitem[Bai et~al.(2022)Bai, Jones, Ndousse, Askell, Chen, DasSarma, Drain, Fort, Ganguli, Henighan, Joseph, Kadavath, Kernion, Conerly, Showk, Elhage, Hatfield{-}Dodds, Hernandez, Hume, Johnston, Kravec, Lovitt, Nanda, Olsson, Amodei, Brown, Clark, McCandlish, Olah, Mann, and Kaplan]{DBLP:journals/corr/abs-2204-05862}
Yuntao Bai, Andy Jones, Kamal Ndousse, Amanda Askell, Anna Chen, Nova DasSarma, Dawn Drain, Stanislav Fort, Deep Ganguli, Tom Henighan, Nicholas Joseph, Saurav Kadavath, Jackson Kernion, Tom Conerly, Sheer~El Showk, Nelson Elhage, Zac Hatfield{-}Dodds, Danny Hernandez, Tristan Hume, Scott Johnston, Shauna Kravec, Liane Lovitt, Neel Nanda, Catherine Olsson, Dario Amodei, Tom~B. Brown, Jack Clark, Sam McCandlish, Chris Olah, Benjamin Mann, and Jared Kaplan.
\newblock Training a helpful and harmless assistant with reinforcement learning from human feedback.
\newblock \emph{CoRR}, abs/2204.05862, 2022.
\newblock \doi{10.48550/ARXIV.2204.05862}.
\newblock URL \url{https://doi.org/10.48550/arXiv.2204.05862}.

\bibitem[Besta et~al.(2024)Besta, Blach, Kubicek, Gerstenberger, Podstawski, Gianinazzi, Gajda, Lehmann, Niewiadomski, Nyczyk, and Hoefler]{DBLP:conf/aaai/BestaBKGPGGLNNH24}
Maciej Besta, Nils Blach, Ales Kubicek, Robert Gerstenberger, Michal Podstawski, Lukas Gianinazzi, Joanna Gajda, Tomasz Lehmann, Hubert Niewiadomski, Piotr Nyczyk, and Torsten Hoefler.
\newblock Graph of thoughts: Solving elaborate problems with large language models.
\newblock In Michael~J. Wooldridge, Jennifer~G. Dy, and Sriraam Natarajan (eds.), \emph{Thirty-Eighth {AAAI} Conference on Artificial Intelligence, {AAAI} 2024, Thirty-Sixth Conference on Innovative Applications of Artificial Intelligence, {IAAI} 2024, Fourteenth Symposium on Educational Advances in Artificial Intelligence, {EAAI} 2014, February 20-27, 2024, Vancouver, Canada}, pp.\  17682--17690. {AAAI} Press, 2024.
\newblock \doi{10.1609/AAAI.V38I16.29720}.
\newblock URL \url{https://doi.org/10.1609/aaai.v38i16.29720}.

\bibitem[Cobbe et~al.(2021)Cobbe, Kosaraju, Bavarian, Chen, Jun, Kaiser, Plappert, Tworek, Hilton, Nakano, Hesse, and Schulman]{DBLP:journals/corr/abs-2110-14168}
Karl Cobbe, Vineet Kosaraju, Mohammad Bavarian, Mark Chen, Heewoo Jun, Lukasz Kaiser, Matthias Plappert, Jerry Tworek, Jacob Hilton, Reiichiro Nakano, Christopher Hesse, and John Schulman.
\newblock Training verifiers to solve math word problems.
\newblock \emph{CoRR}, abs/2110.14168, 2021.
\newblock URL \url{https://arxiv.org/abs/2110.14168}.

\bibitem[DeepSeek-AI et~al.(2025)DeepSeek-AI, Guo, Yang, Zhang, Song, Zhang, Xu, Zhu, Ma, Wang, Bi, Zhang, Yu, Wu, Wu, Gou, Shao, Li, Gao, Liu, Xue, Wang, Wu, Feng, Lu, Zhao, Deng, Zhang, Ruan, Dai, Chen, Ji, Li, Lin, Dai, Luo, Hao, Chen, Li, Zhang, Bao, Xu, Wang, Ding, Xin, Gao, Qu, Li, Guo, Li, Wang, Chen, Yuan, Qiu, Li, Cai, Ni, Liang, Chen, Dong, Hu, Gao, Guan, Huang, Yu, Wang, Zhang, Zhao, Wang, Zhang, Xu, Xia, Zhang, Zhang, Tang, Li, Wang, Li, Tian, Huang, Zhang, Wang, Chen, Du, Ge, Zhang, Pan, Wang, Chen, Jin, Chen, Lu, Zhou, Chen, Ye, Wang, Yu, Zhou, Pan, Li, Zhou, Wu, Ye, Yun, Pei, Sun, Wang, Zeng, Zhao, Liu, Liang, Gao, Yu, Zhang, Xiao, An, Liu, Wang, Chen, Nie, Cheng, Liu, Xie, Liu, Yang, Li, Su, Lin, Li, Jin, Shen, Chen, Sun, Wang, Song, Zhou, Wang, Shan, Li, Wang, Wei, Zhang, Xu, Li, Zhao, Sun, Wang, Yu, Zhang, Shi, Xiong, He, Piao, Wang, Tan, Ma, Liu, Guo, Ou, Wang, Gong, Zou, He, Xiong, Luo, You, Liu, Zhou, Zhu, Xu, Huang, Li, Zheng, Zhu, Ma, Tang, Zha, Yan, Ren, Ren, Sha, Fu, Xu, Xie, Zhang,
  Hao, Ma, Yan, Wu, Gu, Zhu, Liu, Li, Xie, Song, Pan, Huang, Xu, Zhang, and Zhang]{deepseekai2025deepseekr1incentivizingreasoningcapability}
DeepSeek-AI, Daya Guo, Dejian Yang, Haowei Zhang, Junxiao Song, Ruoyu Zhang, Runxin Xu, Qihao Zhu, Shirong Ma, Peiyi Wang, Xiao Bi, Xiaokang Zhang, Xingkai Yu, Yu~Wu, Z.~F. Wu, Zhibin Gou, Zhihong Shao, Zhuoshu Li, Ziyi Gao, Aixin Liu, Bing Xue, Bingxuan Wang, Bochao Wu, Bei Feng, Chengda Lu, Chenggang Zhao, Chengqi Deng, Chenyu Zhang, Chong Ruan, Damai Dai, Deli Chen, Dongjie Ji, Erhang Li, Fangyun Lin, Fucong Dai, Fuli Luo, Guangbo Hao, Guanting Chen, Guowei Li, H.~Zhang, Han Bao, Hanwei Xu, Haocheng Wang, Honghui Ding, Huajian Xin, Huazuo Gao, Hui Qu, Hui Li, Jianzhong Guo, Jiashi Li, Jiawei Wang, Jingchang Chen, Jingyang Yuan, Junjie Qiu, Junlong Li, J.~L. Cai, Jiaqi Ni, Jian Liang, Jin Chen, Kai Dong, Kai Hu, Kaige Gao, Kang Guan, Kexin Huang, Kuai Yu, Lean Wang, Lecong Zhang, Liang Zhao, Litong Wang, Liyue Zhang, Lei Xu, Leyi Xia, Mingchuan Zhang, Minghua Zhang, Minghui Tang, Meng Li, Miaojun Wang, Mingming Li, Ning Tian, Panpan Huang, Peng Zhang, Qiancheng Wang, Qinyu Chen, Qiushi Du, Ruiqi Ge, Ruisong
  Zhang, Ruizhe Pan, Runji Wang, R.~J. Chen, R.~L. Jin, Ruyi Chen, Shanghao Lu, Shangyan Zhou, Shanhuang Chen, Shengfeng Ye, Shiyu Wang, Shuiping Yu, Shunfeng Zhou, Shuting Pan, S.~S. Li, Shuang Zhou, Shaoqing Wu, Shengfeng Ye, Tao Yun, Tian Pei, Tianyu Sun, T.~Wang, Wangding Zeng, Wanjia Zhao, Wen Liu, Wenfeng Liang, Wenjun Gao, Wenqin Yu, Wentao Zhang, W.~L. Xiao, Wei An, Xiaodong Liu, Xiaohan Wang, Xiaokang Chen, Xiaotao Nie, Xin Cheng, Xin Liu, Xin Xie, Xingchao Liu, Xinyu Yang, Xinyuan Li, Xuecheng Su, Xuheng Lin, X.~Q. Li, Xiangyue Jin, Xiaojin Shen, Xiaosha Chen, Xiaowen Sun, Xiaoxiang Wang, Xinnan Song, Xinyi Zhou, Xianzu Wang, Xinxia Shan, Y.~K. Li, Y.~Q. Wang, Y.~X. Wei, Yang Zhang, Yanhong Xu, Yao Li, Yao Zhao, Yaofeng Sun, Yaohui Wang, Yi~Yu, Yichao Zhang, Yifan Shi, Yiliang Xiong, Ying He, Yishi Piao, Yisong Wang, Yixuan Tan, Yiyang Ma, Yiyuan Liu, Yongqiang Guo, Yuan Ou, Yuduan Wang, Yue Gong, Yuheng Zou, Yujia He, Yunfan Xiong, Yuxiang Luo, Yuxiang You, Yuxuan Liu, Yuyang Zhou, Y.~X. Zhu,
  Yanhong Xu, Yanping Huang, Yaohui Li, Yi~Zheng, Yuchen Zhu, Yunxian Ma, Ying Tang, Yukun Zha, Yuting Yan, Z.~Z. Ren, Zehui Ren, Zhangli Sha, Zhe Fu, Zhean Xu, Zhenda Xie, Zhengyan Zhang, Zhewen Hao, Zhicheng Ma, Zhigang Yan, Zhiyu Wu, Zihui Gu, Zijia Zhu, Zijun Liu, Zilin Li, Ziwei Xie, Ziyang Song, Zizheng Pan, Zhen Huang, Zhipeng Xu, Zhongyu Zhang, and Zhen Zhang.
\newblock Deepseek-r1: Incentivizing reasoning capability in llms via reinforcement learning, 2025.
\newblock URL \url{https://arxiv.org/abs/2501.12948}.

\bibitem[Dubey et~al.(2024)Dubey, Jauhri, Pandey, Kadian, Al-Dahle, Letman, Mathur, Schelten, Yang, Fan, et~al.]{dubey2024llama}
Abhimanyu Dubey, Abhinav Jauhri, Abhinav Pandey, Abhishek Kadian, Ahmad Al-Dahle, Aiesha Letman, Akhil Mathur, Alan Schelten, Amy Yang, Angela Fan, et~al.
\newblock The llama 3 herd of models.
\newblock \emph{arXiv preprint arXiv:2407.21783}, 2024.

\bibitem[Fan et~al.(2025)Fan, Li, Sun, and Zhou]{DBLP:journals/corr/abs-2504-06514}
Chenrui Fan, Ming Li, Lichao Sun, and Tianyi Zhou.
\newblock Missing premise exacerbates overthinking: Are reasoning models losing critical thinking skill?
\newblock \emph{CoRR}, abs/2504.06514, 2025.
\newblock \doi{10.48550/ARXIV.2504.06514}.
\newblock URL \url{https://doi.org/10.48550/arXiv.2504.06514}.

\bibitem[Fu et~al.(2025)Fu, Wang, Tian, and Zhao]{DBLP:journals/corr/abs-2508-15260}
Yichao Fu, Xuewei Wang, Yuandong Tian, and Jiawei Zhao.
\newblock Deep think with confidence.
\newblock \emph{CoRR}, abs/2508.15260, 2025.
\newblock \doi{10.48550/ARXIV.2508.15260}.
\newblock URL \url{https://doi.org/10.48550/arXiv.2508.15260}.

\bibitem[Gan et~al.(2024)Gan, Li, Xie, Wen, Purver, and Poesio]{gan2024clarq}
Yujian Gan, Changling Li, Jinxia Xie, Luou Wen, Matthew Purver, and Massimo Poesio.
\newblock Clarq-llm: A benchmark for models clarifying and requesting information in task-oriented dialog.
\newblock \emph{arXiv preprint arXiv:2409.06097}, 2024.

\bibitem[Gao(2025)]{gao2025modeling}
Zhenyu Gao.
\newblock Modeling reasoning as markov decision processes: A theoretical investigation into nlp transformer models.
\newblock 2025.

\bibitem[Hao et~al.(2025)Hao, Li, Yuan, and Li]{hao2025rl}
Qianyue Hao, Sibo Li, Jian Yuan, and Yong Li.
\newblock Rl of thoughts: Navigating llm reasoning with inference-time reinforcement learning.
\newblock \emph{arXiv preprint arXiv:2505.14140}, 2025.

\bibitem[Hu et~al.(2020)Hu, Wen, Wang, Li, and de~Melo]{DBLP:conf/coling/HuWWLM20}
Xiang Hu, Zujie Wen, Yafang Wang, Xiaolong Li, and Gerard de~Melo.
\newblock Interactive question clarification in dialogue via reinforcement learning.
\newblock In Ann Clifton and Courtney Napoles (eds.), \emph{Proceedings of the 28th International Conference on Computational Linguistics, {COLING} 2020 - Industry Track, Online, December 12, 2020}, pp.\  78--89. International Committee on Computational Linguistics, 2020.
\newblock \doi{10.18653/V1/2020.COLING-INDUSTRY.8}.
\newblock URL \url{https://doi.org/10.18653/v1/2020.coling-industry.8}.

\bibitem[Huang et~al.(2025)Huang, Yu, Ma, Zhong, Feng, Wang, Chen, Peng, Feng, Qin, and Liu]{DBLP:journals/tois/HuangYMZFWCPFQL25}
Lei Huang, Weijiang Yu, Weitao Ma, Weihong Zhong, Zhangyin Feng, Haotian Wang, Qianglong Chen, Weihua Peng, Xiaocheng Feng, Bing Qin, and Ting Liu.
\newblock A survey on hallucination in large language models: Principles, taxonomy, challenges, and open questions.
\newblock \emph{{ACM} Trans. Inf. Syst.}, 43\penalty0 (2):\penalty0 42:1--42:55, 2025.
\newblock \doi{10.1145/3703155}.
\newblock URL \url{https://doi.org/10.1145/3703155}.

\bibitem[Jiang et~al.(2025)Jiang, Xiong, Yuan, Xin, Xu, Yue, Zhao, and Yan]{jiang2025pagmultiturnreinforcedllm}
Yuhua Jiang, Yuwen Xiong, Yufeng Yuan, Chao Xin, Wenyuan Xu, Yu~Yue, Qianchuan Zhao, and Lin Yan.
\newblock Pag: Multi-turn reinforced llm self-correction with policy as generative verifier, 2025.
\newblock URL \url{https://arxiv.org/abs/2506.10406}.

\bibitem[Jin et~al.(2025)Jin, Zeng, Yue, Yoon, Arik, Wang, Zamani, and Han]{jin2025search}
Bowen Jin, Hansi Zeng, Zhenrui Yue, Jinsung Yoon, Sercan Arik, Dong Wang, Hamed Zamani, and Jiawei Han.
\newblock Search-r1: Training llms to reason and leverage search engines with reinforcement learning.
\newblock \emph{arXiv preprint arXiv:2503.09516}, 2025.

\bibitem[Kadavath et~al.(2022)Kadavath, Conerly, Askell, Henighan, Drain, Perez, Schiefer, Hatfield-Dodds, DasSarma, Tran-Johnson, et~al.]{kadavath2022language}
Saurav Kadavath, Tom Conerly, Amanda Askell, Tom Henighan, Dawn Drain, Ethan Perez, Nicholas Schiefer, Zac Hatfield-Dodds, Nova DasSarma, Eli Tran-Johnson, et~al.
\newblock Language models (mostly) know what they know.
\newblock \emph{arXiv preprint arXiv:2207.05221}, 2022.

\bibitem[Khalifa et~al.(2025)Khalifa, Agarwal, Logeswaran, Kim, Peng, Lee, Lee, and Wang]{khalifa2025process}
Muhammad Khalifa, Rishabh Agarwal, Lajanugen Logeswaran, Jaekyeom Kim, Hao Peng, Moontae Lee, Honglak Lee, and Lu~Wang.
\newblock Process reward models that think.
\newblock \emph{arXiv preprint arXiv:2504.16828}, 2025.

\bibitem[Kuhn et~al.(2022)Kuhn, Gal, and Farquhar]{kuhn2022clam}
Lorenz Kuhn, Yarin Gal, and Sebastian Farquhar.
\newblock Clam: Selective clarification for ambiguous questions with generative language models.
\newblock \emph{arXiv preprint arXiv:2212.07769}, 2022.

\bibitem[Laban et~al.(2025)Laban, Hayashi, Zhou, and Neville]{DBLP:journals/corr/abs-2505-06120}
Philippe Laban, Hiroaki Hayashi, Yingbo Zhou, and Jennifer Neville.
\newblock Llms get lost in multi-turn conversation.
\newblock \emph{CoRR}, abs/2505.06120, 2025.
\newblock \doi{10.48550/ARXIV.2505.06120}.
\newblock URL \url{https://doi.org/10.48550/arXiv.2505.06120}.

\bibitem[Li et~al.(2023)Li, Patel, Vi{\'e}gas, Pfister, and Wattenberg]{li2023inference}
Kenneth Li, Oam Patel, Fernanda Vi{\'e}gas, Hanspeter Pfister, and Martin Wattenberg.
\newblock Inference-time intervention: Eliciting truthful answers from a language model.
\newblock \emph{Advances in Neural Information Processing Systems}, 36:\penalty0 41451--41530, 2023.

\bibitem[Li et~al.(2025)Li, Dong, Jin, Zhang, Zhou, Zhu, Zhang, and Dou]{li2025search}
Xiaoxi Li, Guanting Dong, Jiajie Jin, Yuyao Zhang, Yujia Zhou, Yutao Zhu, Peitian Zhang, and Zhicheng Dou.
\newblock Search-o1: Agentic search-enhanced large reasoning models.
\newblock \emph{arXiv preprint arXiv:2501.05366}, 2025.

\bibitem[Lightman et~al.(2023)Lightman, Kosaraju, Burda, Edwards, Baker, Lee, Leike, Schulman, Sutskever, and Cobbe]{lightman2023let}
Hunter Lightman, Vineet Kosaraju, Yuri Burda, Harrison Edwards, Bowen Baker, Teddy Lee, Jan Leike, John Schulman, Ilya Sutskever, and Karl Cobbe.
\newblock Let's verify step by step.
\newblock In \emph{The Twelfth International Conference on Learning Representations}, 2023.

\bibitem[Liu et~al.(2025)Liu, Gu, Zhang, Qian, and He]{liu2025reversethinkingenhancesmissing}
Yuxin Liu, Chaojie Gu, Yihang Zhang, Bin Qian, and Shibo He.
\newblock Reverse thinking enhances missing information detection in large language models, 2025.
\newblock URL \url{https://arxiv.org/abs/2512.10273}.

\bibitem[Luo et~al.(2024)Luo, Liu, Liu, Phatale, Lara, Li, Shu, Zhu, Meng, Sun, and Rastogi]{DBLP:journals/corr/abs-2406-06592}
Liangchen Luo, Yinxiao Liu, Rosanne Liu, Samrat Phatale, Harsh Lara, Yunxuan Li, Lei Shu, Yun Zhu, Lei Meng, Jiao Sun, and Abhinav Rastogi.
\newblock Improve mathematical reasoning in language models by automated process supervision.
\newblock \emph{CoRR}, abs/2406.06592, 2024.
\newblock \doi{10.48550/ARXIV.2406.06592}.
\newblock URL \url{https://doi.org/10.48550/arXiv.2406.06592}.

\bibitem[Luo et~al.(2025)Luo, Huang, Li, Meng, Liu, Hu, Feng, and Liu]{luo2025clarifymtbenchbenchmarkingimprovingmultiturn}
Sichun Luo, Yi~Huang, Mukai Li, Shichang Meng, Fengyuan Liu, Zefa Hu, Junlan Feng, and Qi~Liu.
\newblock Clarifymt-bench: Benchmarking and improving multi-turn clarification for conversational large language models, 2025.
\newblock URL \url{https://arxiv.org/abs/2512.21120}.

\bibitem[Madaan et~al.(2023)Madaan, Tandon, Gupta, Hallinan, Gao, Wiegreffe, Alon, Dziri, Prabhumoye, Yang, Gupta, Majumder, Hermann, Welleck, Yazdanbakhsh, and Clark]{DBLP:conf/nips/MadaanTGHGW0DPY23}
Aman Madaan, Niket Tandon, Prakhar Gupta, Skyler Hallinan, Luyu Gao, Sarah Wiegreffe, Uri Alon, Nouha Dziri, Shrimai Prabhumoye, Yiming Yang, Shashank Gupta, Bodhisattwa~Prasad Majumder, Katherine Hermann, Sean Welleck, Amir Yazdanbakhsh, and Peter Clark.
\newblock Self-refine: Iterative refinement with self-feedback.
\newblock In Alice Oh, Tristan Naumann, Amir Globerson, Kate Saenko, Moritz Hardt, and Sergey Levine (eds.), \emph{Advances in Neural Information Processing Systems 36: Annual Conference on Neural Information Processing Systems 2023, NeurIPS 2023, New Orleans, LA, USA, December 10 - 16, 2023}, 2023.
\newblock URL \url{http://papers.nips.cc/paper\_files/paper/2023/hash/91edff07232fb1b55a505a9e9f6c0ff3-Abstract-Conference.html}.

\bibitem[OpenAI(2023)]{DBLP:journals/corr/abs-2303-08774}
OpenAI.
\newblock {GPT-4} technical report.
\newblock \emph{CoRR}, abs/2303.08774, 2023.
\newblock \doi{10.48550/ARXIV.2303.08774}.
\newblock URL \url{https://doi.org/10.48550/arXiv.2303.08774}.

\bibitem[Ouyang et~al.(2022)Ouyang, Wu, Jiang, Almeida, Wainwright, Mishkin, Zhang, Agarwal, Slama, Ray, et~al.]{ouyang2022training}
Long Ouyang, Jeffrey Wu, Xu~Jiang, Diogo Almeida, Carroll Wainwright, Pamela Mishkin, Chong Zhang, Sandhini Agarwal, Katarina Slama, Alex Ray, et~al.
\newblock Training language models to follow instructions with human feedback.
\newblock \emph{Advances in neural information processing systems}, 35:\penalty0 27730--27744, 2022.

\bibitem[Rita et~al.(2024)Rita, Strub, Chaabouni, Michel, Dupoux, and Pietquin]{rita2024countering}
Mathieu Rita, Florian Strub, Rahma Chaabouni, Paul Michel, Emmanuel Dupoux, and Olivier Pietquin.
\newblock Countering reward over-optimization in llm with demonstration-guided reinforcement learning.
\newblock \emph{arXiv preprint arXiv:2404.19409}, 2024.

\bibitem[Schick et~al.(2023)Schick, Dwivedi{-}Yu, Dess{\`{\i}}, Raileanu, Lomeli, Hambro, Zettlemoyer, Cancedda, and Scialom]{DBLP:conf/nips/SchickDDRLHZCS23}
Timo Schick, Jane Dwivedi{-}Yu, Roberto Dess{\`{\i}}, Roberta Raileanu, Maria Lomeli, Eric Hambro, Luke Zettlemoyer, Nicola Cancedda, and Thomas Scialom.
\newblock Toolformer: Language models can teach themselves to use tools.
\newblock In Alice Oh, Tristan Naumann, Amir Globerson, Kate Saenko, Moritz Hardt, and Sergey Levine (eds.), \emph{Advances in Neural Information Processing Systems 36: Annual Conference on Neural Information Processing Systems 2023, NeurIPS 2023, New Orleans, LA, USA, December 10 - 16, 2023}, 2023.
\newblock URL \url{http://papers.nips.cc/paper\_files/paper/2023/hash/d842425e4bf79ba039352da0f658a906-Abstract-Conference.html}.

\bibitem[Schulman et~al.(2017)Schulman, Wolski, Dhariwal, Radford, and Klimov]{DBLP:journals/corr/SchulmanWDRK17}
John Schulman, Filip Wolski, Prafulla Dhariwal, Alec Radford, and Oleg Klimov.
\newblock Proximal policy optimization algorithms.
\newblock \emph{CoRR}, abs/1707.06347, 2017.
\newblock URL \url{http://arxiv.org/abs/1707.06347}.

\bibitem[Shao et~al.(2024)Shao, Wang, Zhu, Xu, Song, Zhang, Li, Wu, and Guo]{DBLP:journals/corr/abs-2402-03300}
Zhihong Shao, Peiyi Wang, Qihao Zhu, Runxin Xu, Junxiao Song, Mingchuan Zhang, Y.~K. Li, Y.~Wu, and Daya Guo.
\newblock Deepseekmath: Pushing the limits of mathematical reasoning in open language models.
\newblock \emph{CoRR}, abs/2402.03300, 2024.
\newblock \doi{10.48550/ARXIV.2402.03300}.
\newblock URL \url{https://doi.org/10.48550/arXiv.2402.03300}.

\bibitem[Sharma et~al.(2023)Sharma, Tong, Korbak, Duvenaud, Askell, Bowman, Cheng, Durmus, Hatfield-Dodds, Johnston, et~al.]{sharma2023towards}
Mrinank Sharma, Meg Tong, Tomasz Korbak, David Duvenaud, Amanda Askell, Samuel~R Bowman, Newton Cheng, Esin Durmus, Zac Hatfield-Dodds, Scott~R Johnston, et~al.
\newblock Towards understanding sycophancy in language models.
\newblock \emph{arXiv preprint arXiv:2310.13548}, 2023.

\bibitem[She et~al.(2025)She, Liu, Liu, Chen, Huang, and Huang]{DBLP:journals/corr/abs-2503-21295}
Shuaijie She, Junxiao Liu, Yifeng Liu, Jiajun Chen, Xin Huang, and Shujian Huang.
\newblock {R-PRM:} reasoning-driven process reward modeling.
\newblock \emph{CoRR}, abs/2503.21295, 2025.
\newblock \doi{10.48550/ARXIV.2503.21295}.
\newblock URL \url{https://doi.org/10.48550/arXiv.2503.21295}.

\bibitem[Sheng et~al.(2025)Sheng, Zhang, Ye, Wu, Zhang, Zhang, Peng, Lin, and Wu]{DBLP:conf/eurosys/ShengZYWZZPL025}
Guangming Sheng, Chi Zhang, Zilingfeng Ye, Xibin Wu, Wang Zhang, Ru~Zhang, Yanghua Peng, Haibin Lin, and Chuan Wu.
\newblock Hybridflow: {A} flexible and efficient {RLHF} framework.
\newblock In \emph{Proceedings of the Twentieth European Conference on Computer Systems, EuroSys 2025, Rotterdam, The Netherlands, 30 March 2025 - 3 April 2025}, pp.\  1279--1297. {ACM}, 2025.
\newblock \doi{10.1145/3689031.3696075}.
\newblock URL \url{https://doi.org/10.1145/3689031.3696075}.

\bibitem[Shinn et~al.(2023)Shinn, Cassano, Gopinath, Narasimhan, and Yao]{shinn2023reflexion}
Noah Shinn, Federico Cassano, Ashwin Gopinath, Karthik Narasimhan, and Shunyu Yao.
\newblock Reflexion: Language agents with verbal reinforcement learning.
\newblock \emph{Advances in Neural Information Processing Systems}, 36:\penalty0 8634--8652, 2023.

\bibitem[Sui et~al.(2025)Sui, Chuang, Wang, Zhang, Zhang, Yuan, Liu, Wen, Zhong, Zou, et~al.]{sui2025stop}
Yang Sui, Yu-Neng Chuang, Guanchu Wang, Jiamu Zhang, Tianyi Zhang, Jiayi Yuan, Hongyi Liu, Andrew Wen, Shaochen Zhong, Na~Zou, et~al.
\newblock Stop overthinking: A survey on efficient reasoning for large language models.
\newblock \emph{arXiv preprint arXiv:2503.16419}, 2025.

\bibitem[Wang et~al.(2024)Wang, Li, Shao, Xu, Dai, Li, Chen, Wu, and Sui]{wang2024math}
Peiyi Wang, Lei Li, Zhihong Shao, Runxin Xu, Damai Dai, Yifei Li, Deli Chen, Yu~Wu, and Zhifang Sui.
\newblock Math-shepherd: Verify and reinforce llms step-by-step without human annotations.
\newblock In \emph{Proceedings of the 62nd Annual Meeting of the Association for Computational Linguistics (Volume 1: Long Papers)}, pp.\  9426--9439, 2024.

\bibitem[Wang et~al.(2023{\natexlab{a}})Wang, Wei, Schuurmans, Le, Chi, Narang, Chowdhery, and Zhou]{DBLP:conf/iclr/0002WSLCNCZ23}
Xuezhi Wang, Jason Wei, Dale Schuurmans, Quoc~V. Le, Ed~H. Chi, Sharan Narang, Aakanksha Chowdhery, and Denny Zhou.
\newblock Self-consistency improves chain of thought reasoning in language models.
\newblock In \emph{The Eleventh International Conference on Learning Representations, {ICLR} 2023, Kigali, Rwanda, May 1-5, 2023}. OpenReview.net, 2023{\natexlab{a}}.
\newblock URL \url{https://openreview.net/forum?id=1PL1NIMMrw}.

\bibitem[Wang et~al.(2025{\natexlab{a}})Wang, Yu, Liang, and He]{wang2025comprehensive}
Yanbo Wang, Yongcan Yu, Jian Liang, and Ran He.
\newblock A comprehensive survey on trustworthiness in reasoning with large language models.
\newblock \emph{arXiv preprint arXiv:2509.03871}, 2025{\natexlab{a}}.

\bibitem[Wang et~al.(2023{\natexlab{b}})Wang, Zhang, Yang, Shi, Zhou, Hao, Xiong, Li, Sim, Chen, et~al.]{wang2023interactive}
Zekun Wang, Ge~Zhang, Kexin Yang, Ning Shi, Wangchunshu Zhou, Shaochun Hao, Guangzheng Xiong, Yizhi Li, Mong~Yuan Sim, Xiuying Chen, et~al.
\newblock Interactive natural language processing.
\newblock \emph{arXiv preprint arXiv:2305.13246}, 2023{\natexlab{b}}.

\bibitem[Wang et~al.(2025{\natexlab{b}})Wang, Wang, Wang, Zhang, Li, Yang, Jin, Yu, Nguyen, Liu, Gottlieb, Lu, Cho, Wu, Fei{-}Fei, Wang, Choi, and Li]{DBLP:journals/corr/abs-2504-20073}
Zihan Wang, Kangrui Wang, Qineng Wang, Pingyue Zhang, Linjie Li, Zhengyuan Yang, Xing Jin, Kefan Yu, Minh~Nhat Nguyen, Licheng Liu, Eli Gottlieb, Yiping Lu, Kyunghyun Cho, Jiajun Wu, Li~Fei{-}Fei, Lijuan Wang, Yejin Choi, and Manling Li.
\newblock {RAGEN:} understanding self-evolution in {LLM} agents via multi-turn reinforcement learning.
\newblock \emph{CoRR}, abs/2504.20073, 2025{\natexlab{b}}.
\newblock \doi{10.48550/ARXIV.2504.20073}.
\newblock URL \url{https://doi.org/10.48550/arXiv.2504.20073}.

\bibitem[Wei et~al.(2022)Wei, Wang, Schuurmans, Bosma, Ichter, Xia, Chi, Le, and Zhou]{DBLP:conf/nips/Wei0SBIXCLZ22}
Jason Wei, Xuezhi Wang, Dale Schuurmans, Maarten Bosma, Brian Ichter, Fei Xia, Ed~H. Chi, Quoc~V. Le, and Denny Zhou.
\newblock Chain-of-thought prompting elicits reasoning in large language models.
\newblock In Sanmi Koyejo, S.~Mohamed, A.~Agarwal, Danielle Belgrave, K.~Cho, and A.~Oh (eds.), \emph{Advances in Neural Information Processing Systems 35: Annual Conference on Neural Information Processing Systems 2022, NeurIPS 2022, New Orleans, LA, USA, November 28 - December 9, 2022}, 2022.
\newblock URL \url{http://papers.nips.cc/paper\_files/paper/2022/hash/9d5609613524ecf4f15af0f7b31abca4-Abstract-Conference.html}.

\bibitem[Xiong et~al.(2024)Xiong, Hu, Lu, Li, Fu, He, and Hooi]{xiong2024can}
Miao Xiong, Zhiyuan Hu, Xinyang Lu, Yifei Li, Jie Fu, Junxian He, and Bryan Hooi.
\newblock Can llms express their uncertainty.
\newblock \emph{An Empirical Evaluation of Confidence Elicitation in LLMs. arXiv}, 2024.

\bibitem[Yan et~al.(2025)Yan, Shen, Liu, Jiang, Xu, Zhang, Shao, and Zhuang]{yan2025mathfimer}
Yuchen Yan, Yongliang Shen, Yang Liu, Jin Jiang, Xin Xu, Mengdi Zhang, Jian Shao, and Yueting Zhuang.
\newblock Mathfimer: Enhancing mathematical reasoning by expanding reasoning steps through fill-in-the-middle task, 2025.
\newblock URL \url{https://arxiv.org/abs/2502.11684}.

\bibitem[Yang et~al.(2024)Yang, Yang, Zhang, Hui, Zheng, Yu, Li, Liu, Huang, Wei, Lin, Yang, Tu, Zhang, Yang, Yang, Zhou, Lin, Dang, Lu, Bao, Yang, Yu, Li, Xue, Zhang, Zhu, Men, Lin, Li, Xia, Ren, Ren, Fan, Su, Zhang, Wan, Liu, Cui, Zhang, and Qiu]{DBLP:journals/corr/abs-2412-15115}
An~Yang, Baosong Yang, Beichen Zhang, Binyuan Hui, Bo~Zheng, Bowen Yu, Chengyuan Li, Dayiheng Liu, Fei Huang, Haoran Wei, Huan Lin, Jian Yang, Jianhong Tu, Jianwei Zhang, Jianxin Yang, Jiaxi Yang, Jingren Zhou, Junyang Lin, Kai Dang, Keming Lu, Keqin Bao, Kexin Yang, Le~Yu, Mei Li, Mingfeng Xue, Pei Zhang, Qin Zhu, Rui Men, Runji Lin, Tianhao Li, Tingyu Xia, Xingzhang Ren, Xuancheng Ren, Yang Fan, Yang Su, Yichang Zhang, Yu~Wan, Yuqiong Liu, Zeyu Cui, Zhenru Zhang, and Zihan Qiu.
\newblock Qwen2.5 technical report.
\newblock \emph{CoRR}, abs/2412.15115, 2024.
\newblock \doi{10.48550/ARXIV.2412.15115}.
\newblock URL \url{https://doi.org/10.48550/arXiv.2412.15115}.

\bibitem[Yang et~al.(2025)Yang, Li, Yang, Zhang, Hui, Zheng, Yu, Gao, Huang, Lv, Zheng, Liu, Zhou, Huang, Hu, Ge, Wei, Lin, Tang, Yang, Tu, Zhang, Yang, Yang, Zhou, Lin, Dang, Bao, Yang, Yu, Deng, Li, Xue, Li, Zhang, Wang, Zhu, Men, Gao, Liu, Luo, Li, Tang, Yin, Ren, Wang, Zhang, Ren, Fan, Su, Zhang, Zhang, Wan, Liu, Wang, Cui, Zhang, Zhou, and Qiu]{DBLP:journals/corr/abs-2505-09388}
An~Yang, Anfeng Li, Baosong Yang, Beichen Zhang, Binyuan Hui, Bo~Zheng, Bowen Yu, Chang Gao, Chengen Huang, Chenxu Lv, Chujie Zheng, Dayiheng Liu, Fan Zhou, Fei Huang, Feng Hu, Hao Ge, Haoran Wei, Huan Lin, Jialong Tang, Jian Yang, Jianhong Tu, Jianwei Zhang, Jian Yang, Jiaxi Yang, Jingren Zhou, Junyang Lin, Kai Dang, Keqin Bao, Kexin Yang, Le~Yu, Lianghao Deng, Mei Li, Mingfeng Xue, Mingze Li, Pei Zhang, Peng Wang, Qin Zhu, Rui Men, Ruize Gao, Shixuan Liu, Shuang Luo, Tianhao Li, Tianyi Tang, Wenbiao Yin, Xingzhang Ren, Xinyu Wang, Xinyu Zhang, Xuancheng Ren, Yang Fan, Yang Su, Yichang Zhang, Yinger Zhang, Yu~Wan, Yuqiong Liu, Zekun Wang, Zeyu Cui, Zhenru Zhang, Zhipeng Zhou, and Zihan Qiu.
\newblock Qwen3 technical report.
\newblock \emph{CoRR}, abs/2505.09388, 2025.
\newblock \doi{10.48550/ARXIV.2505.09388}.
\newblock URL \url{https://doi.org/10.48550/arXiv.2505.09388}.

\bibitem[Yang et~al.(2018)Yang, Qi, Zhang, Bengio, Cohen, Salakhutdinov, and Manning]{DBLP:conf/emnlp/Yang0ZBCSM18}
Zhilin Yang, Peng Qi, Saizheng Zhang, Yoshua Bengio, William~W. Cohen, Ruslan Salakhutdinov, and Christopher~D. Manning.
\newblock Hotpotqa: {A} dataset for diverse, explainable multi-hop question answering.
\newblock In Ellen Riloff, David Chiang, Julia Hockenmaier, and Jun'ichi Tsujii (eds.), \emph{Proceedings of the 2018 Conference on Empirical Methods in Natural Language Processing, Brussels, Belgium, October 31 - November 4, 2018}, pp.\  2369--2380. Association for Computational Linguistics, 2018.
\newblock \doi{10.18653/V1/D18-1259}.
\newblock URL \url{https://doi.org/10.18653/v1/d18-1259}.

\bibitem[Yao et~al.(2023{\natexlab{a}})Yao, Yu, Zhao, Shafran, Griffiths, Cao, and Narasimhan]{DBLP:conf/nips/YaoYZS00N23}
Shunyu Yao, Dian Yu, Jeffrey Zhao, Izhak Shafran, Tom Griffiths, Yuan Cao, and Karthik Narasimhan.
\newblock Tree of thoughts: Deliberate problem solving with large language models.
\newblock In Alice Oh, Tristan Naumann, Amir Globerson, Kate Saenko, Moritz Hardt, and Sergey Levine (eds.), \emph{Advances in Neural Information Processing Systems 36: Annual Conference on Neural Information Processing Systems 2023, NeurIPS 2023, New Orleans, LA, USA, December 10 - 16, 2023}, 2023{\natexlab{a}}.
\newblock URL \url{http://papers.nips.cc/paper\_files/paper/2023/hash/271db9922b8d1f4dd7aaef84ed5ac703-Abstract-Conference.html}.

\bibitem[Yao et~al.(2023{\natexlab{b}})Yao, Zhao, Yu, Du, Shafran, Narasimhan, and Cao]{DBLP:conf/iclr/YaoZYDSN023}
Shunyu Yao, Jeffrey Zhao, Dian Yu, Nan Du, Izhak Shafran, Karthik~R. Narasimhan, and Yuan Cao.
\newblock React: Synergizing reasoning and acting in language models.
\newblock In \emph{The Eleventh International Conference on Learning Representations, {ICLR} 2023, Kigali, Rwanda, May 1-5, 2023}. OpenReview.net, 2023{\natexlab{b}}.
\newblock URL \url{https://openreview.net/forum?id=WE\_vluYUL-X}.

\bibitem[Yu et~al.(2024)Yu, Jiang, Shi, Yu, Liu, Zhang, Kwok, Li, Weller, and Liu]{DBLP:conf/iclr/YuJSYLZKLWL24}
Longhui Yu, Weisen Jiang, Han Shi, Jincheng Yu, Zhengying Liu, Yu~Zhang, James~T. Kwok, Zhenguo Li, Adrian Weller, and Weiyang Liu.
\newblock Metamath: Bootstrap your own mathematical questions for large language models.
\newblock In \emph{The Twelfth International Conference on Learning Representations, {ICLR} 2024, Vienna, Austria, May 7-11, 2024}. OpenReview.net, 2024.
\newblock URL \url{https://openreview.net/forum?id=N8N0hgNDRt}.

\bibitem[Zhang et~al.(2024)Zhang, Knox, and Choi]{zhang2024modeling}
Michael~JQ Zhang, W~Bradley Knox, and Eunsol Choi.
\newblock Modeling future conversation turns to teach llms to ask clarifying questions.
\newblock \emph{arXiv preprint arXiv:2410.13788}, 2024.

\bibitem[Zhang et~al.(2025)Zhang, Shen, Zheng, Wu, Zhang, Yan, Peng, Wang, and Lu]{DBLP:journals/corr/abs-2503-01940}
Xuan Zhang, Yongliang Shen, Zhe Zheng, Linjuan Wu, Wenqi Zhang, Yuchen Yan, Qiuying Peng, Jun Wang, and Weiming Lu.
\newblock Asktoact: Enhancing llms tool use via self-correcting clarification.
\newblock \emph{CoRR}, abs/2503.01940, 2025.
\newblock \doi{10.48550/ARXIV.2503.01940}.
\newblock URL \url{https://doi.org/10.48550/arXiv.2503.01940}.

\bibitem[Zhou et~al.(2024)Zhou, Zanette, Pan, Levine, and Kumar]{DBLP:conf/icml/ZhouZPLK24}
Yifei Zhou, Andrea Zanette, Jiayi Pan, Sergey Levine, and Aviral Kumar.
\newblock Archer: Training language model agents via hierarchical multi-turn {RL}.
\newblock In \emph{Forty-first International Conference on Machine Learning, {ICML} 2024, Vienna, Austria, July 21-27, 2024}. OpenReview.net, 2024.
\newblock URL \url{https://openreview.net/forum?id=b6rA0kAHT1}.

\end{thebibliography}
\bibliographystyle{main}
\appendix
\section{Algorithm pipeline}
\begin{algorithm}[H]
\caption{GRIL Training Framework}
\label{alg:gril}
\SetKwInOut{Input}{Input}
\SetKwComment{Comment}{$\triangleright$ }{}
\Input{Dataset $\mathcal{D} = \mathcal{D}_{inc} \cup \mathcal{D}_{comp}$, Initial Policy $\pi_\theta$, Reference Policy $\pi_{\text{ref}}$, Max turns $K$, Hyperparameters $\gamma_d, \alpha, \beta_{\text{KL}}, \lambda$}

\For{$\text{iteration} = 1, \dots, I$}{
    Sample batch of queries $Q \sim \mathcal{D}$\;
    \ForEach{query $q \in Q$}{
        Initialize state $s_0 \leftarrow (q)$, turn $t \leftarrow 0$\;
        
        \Comment{Stage 1: Clarify and Pause}
        \While{$t < K$ \textbf{and} not terminated}{
            Generate inferential action $a_t \sim \pi_\theta(\cdot \mid s_t)$\;
            
            \If{$q \in \mathcal{D}_{inc}$}{
                \eIf{$a_t = a_{\text{clarify}}$}{
                    Obtain missing premise $p_q$ from Environment\;
                    $R_{\text{detect}} \leftarrow \gamma_d^t \cdot r_{\text{base}}$\;
                    $s_{t+1} \leftarrow s_t \oplus a_t \oplus p_q$\;
                    
                    \Comment{Stage 2: Grounded Reasoning}
                    Generate solution $a_{\text{solve}} \sim \pi_\theta(\cdot \mid s_{t+1})$\;
                    $R_{\text{solve}} \leftarrow r_{\text{correct}} \cdot \mathbb{I}[a_{\text{solve}} \text{ is correct}]$\;
                    Terminate episode\;
                }{
                    Receive negative feedback $\text{[neg.]}$ from Environment\;
                    $s_{t+1} \leftarrow s_t \oplus a_t \oplus \text{[neg.]}$\;
                }
            }
            \ElseIf{$q \in \mathcal{D}_{comp}$}{
                \eIf{$a_t = a_{\text{solve}}$}{
                    $R_{\text{comp}} \leftarrow r_{\text{correct}} \cdot \mathbb{I}[a_t \text{ is correct}]$\;
                    Terminate episode\;
                }{
                    Receive unnecessary feedback $\text{[unc.]}$ from Environment\;
                    Apply penalty $R_{\text{comp}} \leftarrow -\lambda \cdot \mathbb{I}[\text{unc.}]$\;
                    $s_{t+1} \leftarrow s_t \oplus a_t \oplus \text{[unc.]}$\;
                }
            }
            $t \leftarrow t + 1$\;
        }
        
        \Comment{Reward Assignment}
        Compute trajectory reward $R(q, \theta)$ by combining $R_{\text{detect}}$, $R_{\text{solve}}$, and $R_{\text{comp}}$ (Eq. 7)\;
    }
    Update policy $\pi_\theta$ by maximizing the PPO objective with KL penalty $-\beta_{\text{KL}} D_{\text{KL}}(\pi_\theta \parallel \pi_{\text{ref}})$\;
}
\end{algorithm}

\section{Appendix}
\subsection{Data construction}
Here give the algorithm to generate training data and test data.
This algorithm constructs supervision for \emph{missing-information detection} via \textbf{premise masking}. For each problem $Q \in \mathcal{D}$, it segments the text into sentences $S=\{s_1,\dots,s_n\}$ and filters candidate premise sentences
$S_{\text{num}}=\{\, s\in S \mid \texttt{contains\_digit}(s)\land \neg \texttt{is\_question}(s)\,\}$.
It then samples a sentence $s_{\text{mask}}\sim \mathrm{Unif}(S_{\text{num}})$, removes it to form $Q_{\text{masked}}=S\setminus\{s_{\text{mask}}\}$, and queries a pretrained LLM to predict whether $Q_{\text{masked}}$ is \texttt{Solvable} or \texttt{Unsolvable}.
If $Q_{\text{masked}}$ is \texttt{Unsolvable}, $s_{\text{mask}}$ is labeled \emph{essential} and the pair $(Q_{\text{masked}}, s_{\text{mask}})$ is added as a missing-premise example; otherwise $s_{\text{mask}}$ is labeled \emph{redundant}.
The output is a labeled dataset of \emph{essential} vs.\ \emph{redundant} premises.
\begin{algorithm}[htbp]
\caption{Missing Information Detection via Premise Masking}
\label{alg:premise_masking}
\KwIn{Mathematical dataset $\mathcal{D}$; sentence tokenizer; pretrained LLM}
\KwOut{Labeled dataset of essential and redundant premises}

\ForEach{$Q \in \mathcal{D}$}{
  Segment $Q$ into sentences $S=\{s_1,\dots,s_n\}$\;
  $S_{\text{num}} \leftarrow \{\, s\in S \mid \text{contains\_digit}(s)\land \neg\text{is\_question}(s)\,\}$\;
  \If{$S_{\text{num}}=\emptyset$}{
    \textbf{continue}\;
  }
  Sample $s_{\text{mask}} \sim \text{Unif}(S_{\text{num}})$\;
  $Q_{\text{masked}} \leftarrow S \setminus \{s_{\text{mask}}\}$\;
  $V \leftarrow \text{LLM}(\texttt{Solvable?}, Q_{\text{masked}})$\;
  \If{$V=\textit{Unsolvable}$}{
    Label $s_{\text{mask}}$ as \textit{essential}\;
    $\mathcal{D}_{\text{neg}} \leftarrow \mathcal{D}_{\text{neg}} \cup \{(Q_{\text{masked}}, s_{\text{mask}})\}$\;
  }\Else{
    Label $s_{\text{mask}}$ as \textit{redundant}\;
  }
}
\Return Labeled dataset\;
\end{algorithm}
\subsection{Param ablation}
\subsubsection{K turns ablation}
\begin{figure}[htbp]
    \centering
    \includegraphics[width=0.8\columnwidth]{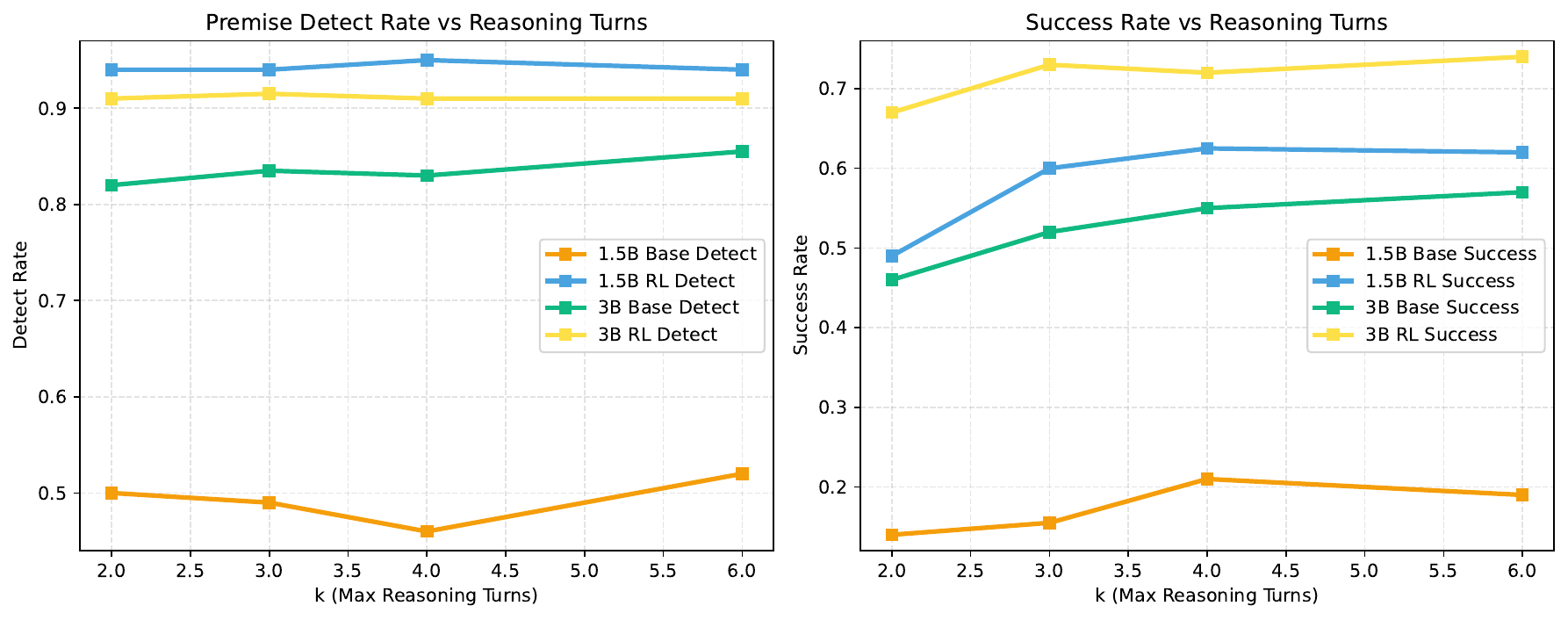} 
    \caption{Performance changes via deifferent iteration turns }
    \label{fig:ablation_one}
\end{figure}

We evaluate premise detection and success rate under varying reasoning turn limits $k$. 
The base model shows unstable performance at small $k$ and improves gradually as $k$ increases. 
In contrast, GRIL consistently improves robustness and performance across all $k$. 
Smaller RL-trained models achieve high premise detection even with limited turns, while larger models reach near-saturated performance within a few turns, indicating that RL promotes early detection and efficient reasoning.
As shown in Figure \ref{fig:ablation_one}, we evaluate premise detection and success rate under varying reasoning turn limits $k$. 
The base model shows unstable performance at small $k$, whereas GRIL consistently improves robustness and 
achieves near-saturated performance with fewer turns.
\subsubsection{Reward weight analysis}
\begin{table}[htbp]
\centering
\caption{Sensitivity analysis of stage weight combinations $(k_1, k_2)$ in Multi-turn RL.}
\label{tab:weight_sensitivity}
\begin{tabular}{c|c|c}
\hline
$(k_1, k_2)$ & Success Rate & Premise Detect \\
\hline
$(0.2, 0.8)$ & 0.624 & 0.918 \\
$(0.3, 0.7)$ & 0.618 & 0.908 \\
$(0.4, 0.6)$ & 0.627 & 0.931 \\
\hline
\end{tabular}
\end{table}
We conduct a parameter sensitivity study to examine the impact of stage weight combinations $(k_1, k_2)$ in GRIL
As shown in Table~\ref{tab:weight_sensitivity}, the Success Rate remains largely stable as $k_1$ increases from 0.2 to 0.4.
In contrast, Premise Detect shows a slight improvement when a larger weight is assigned to the first stage,
indicating that emphasizing the early stage facilitates earlier detection of missing premises without degrading overall problem-solving performance.
These results suggest that GRIL is robust to stage weight variations and maintains effective reasoning and premise awareness across different configurations.
\subsubsection{Data ratio analysis}
\begin{table}[htbp]
\centering
\setlength{\tabcolsep}{12pt}
\begin{tabular}{c|c|c}
\hline
Ratio & Success Rate & Premise Detect \\
\hline
$3:7$ & 0.650 & 0.873 \\
$4:6$ & 0.601 & 0.973 \\
$5:5$ & 0.622 & 0.955 \\
\hline
\end{tabular}
\caption{Effect of data ratio between incomplete-premise and complete queries.}
\label{tab:data_ratio}
\end{table}
We study the impact of different data ratios between incomplete-premise queries and complete queries.
As shown in Table~\ref{tab:data_ratio}, the model achieves competitive Success Rate and Premise Detect
even with a smaller proportion of incomplete-premise data (e.g., $3{:}7$).
As the ratio increases to $4{:}6$ or $5{:}5$, Premise Detect improves, while Success Rate remains relatively stable.
These results indicate that GRIL is robust to the amount of incomplete-premise data
and can effectively learn premise detection and problem solving even when such data is limited.
\subsubsection{Time decay analysis}
\begin{table}[htbp]
    \centering
    \small
    \setlength{\tabcolsep}{10pt} %
    \renewcommand{\arraystretch}{1.05}
    \label{tab:decay_gamma}
    \begin{tabular}{@{}c c c c@{}}
        \toprule
        $\gamma$ & Success Rate & Premise Detect & \makecell{Response\\Length} \\
        \midrule
        0.5 & 0.617 & 0.936 & 472 \\
        1.0 & 0.659 & 0.894 & 838 \\
        \bottomrule
    \end{tabular}
    \caption{Effect of the decay factor $\gamma$ in the first-stage policy.}
    \label{tab:decay_gamma}
\end{table}
To examine the effect of the decay factor in the first-stage policy, we compare $\gamma=0.5$ and $\gamma=1.0$.
As shown in Table~\ref{tab:decay_gamma}, a stronger decay ($\gamma=0.5$) significantly improves Premise Detect,
indicating higher sensitivity to missing premises and a stronger tendency to initiate clarification early.
However, this aggressive querying behavior also leads to a slight drop in Success Rate,
suggesting a more conservative strategy that may prematurely abandon complex problems.
In addition, the stronger decay results in shorter average responses,
while the non-decayed setting ($\gamma=1.0$) produces longer responses due to more extended reasoning trajectories.
\subsection{Baseline and Test Data}
\subsubsection{Baselines}
\label{sec:baselines}

We design three categories of baseline models to systematically analyze the roles of prompting, supervised fine-tuning (SFT), and reinforcement learning (RL) in handling missing-premise scenarios.

\paragraph{Base Model.}
The base model refers to a pretrained language model without any task-specific training.
It is evaluated on all benchmarks using a standard zero-shot reasoning prompt, without explicitly informing the model that the input question may contain missing premises.
This baseline measures the model's intrinsic ability to handle incomplete information relying solely on knowledge acquired during pretraining.

\paragraph{Prompt-based Model.}
The prompt-based model shares the same pretrained backbone as the base model, but augments the input with an explicit task instruction.
Specifically, the prompt informs the model that the problem may contain missing or incomplete premises and requires the model to first assess solvability before attempting to produce a solution.
No parameter updates are performed for this model, allowing us to isolate and evaluate the performance gains introduced purely by explicit task prompting.

\begin{figure}[htbp]
    \centering
    \includegraphics[width=0.8\columnwidth]{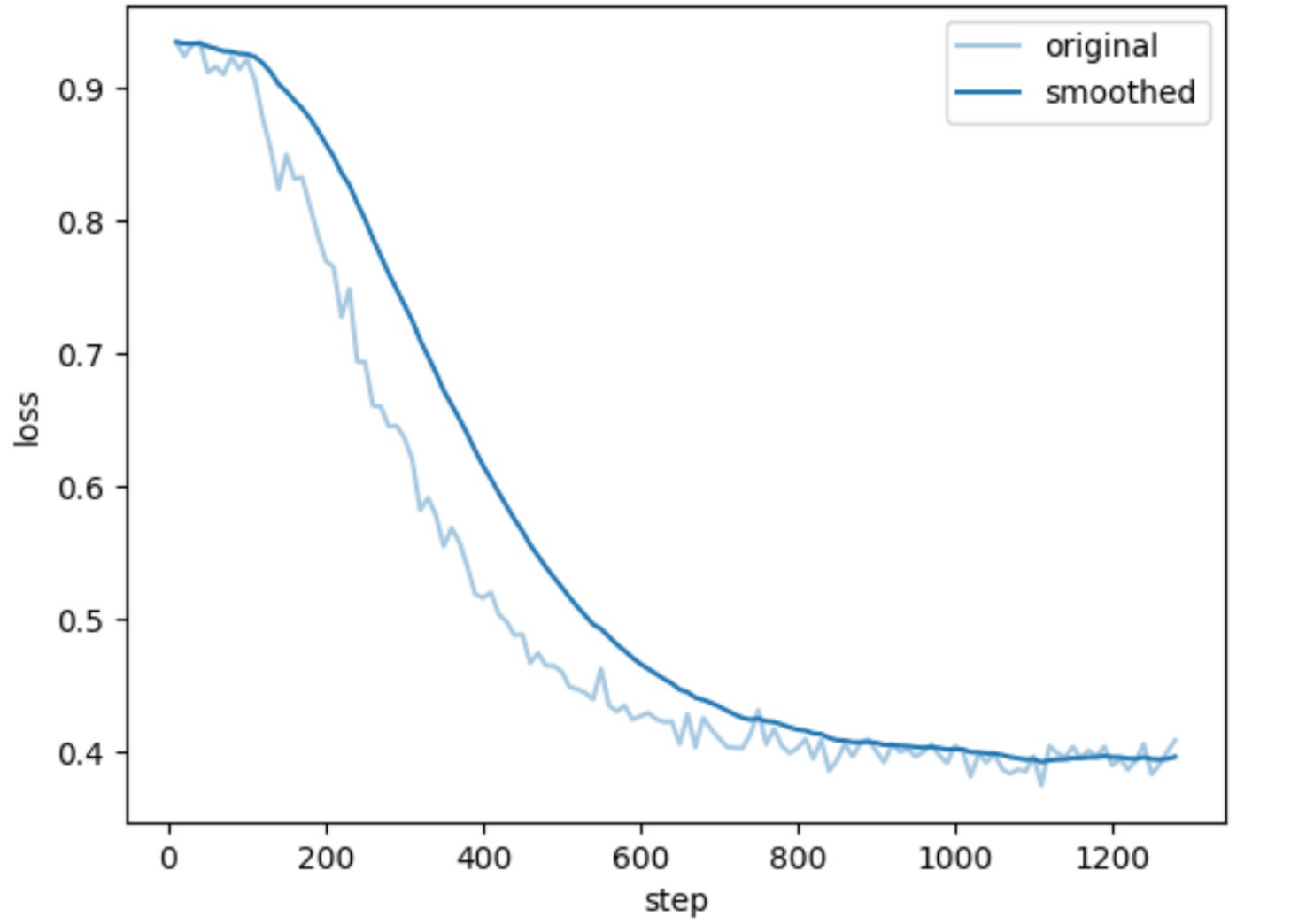} 
    \caption{Example loss plot of SFT Methods }
    \label{fig:sft_loss}
\end{figure}

\paragraph{Supervised Fine-Tuned (SFT) Model.}
The SFT model is trained on the same data source as the reinforcement learning stage, but optimized using supervised learning.
We construct high-quality, multi-turn missing-premise dialogue data using GPT-4o-mini.
Each dialogue trajectory typically consists of:
(1) an initial question with insufficient information,
(2) intermediate premise detection or clarification steps, and
(3) a final outcome, such as requesting additional information or determining that the problem is unsolvable under the current conditions.
The model is fine-tuned by imitating these trajectories, serving as a strong supervised baseline aligned with the RL model in terms of data distribution.Example of training loss are shown in Figure ~\ref{fig:sft_loss}

\subsubsection{Evaluation Datasets}
\label{sec:datasets}

We evaluate all models on a diverse set of benchmark datasets, including both our constructed missing-premise datasets and publicly available benchmarks used in prior work.

\paragraph{GSM8K-Insufficient.}
GSM8K\_Insufficient is constructed from the GSM8K test set by systematically removing essential premises from the original problems, rendering them unsolvable without additional information.
This dataset is designed to evaluate a model's ability to detect missing premises and avoid overconfident reasoning under incomplete inputs.

\paragraph{MetaMath-Insufficient.}
MetaMath\_Insufficient is constructed using the same procedure as GSM8K\_Insufficient, but based on the MetaMath dataset.
Compared to GSM8K, this dataset contains more diverse mathematical structures and longer reasoning chains, posing greater challenges for missing-premise detection in complex scenarios.

\paragraph{Public Missing-Premise Benchmarks.}
In addition, we evaluate our models on publicly available benchmarks from prior work, including GSM8K, SVAMP, and Formula.
These datasets contain manually or semi-automatically constructed missing-premise problems and have been widely used to assess model robustness under incomplete inputs.
Evaluating on these benchmarks ensures the comparability of our approach with existing methods.

\subsection{System prompt example}
We provide system prompt example for better detail, just displayed in Table ~\ref{tab:sys-prompt}, all prompt stays the same during traning and test.
\begin{table}[htbp]
\centering
\begin{tcolorbox}[
    colback=white,
    colframe=black,
    boxrule=1pt,
    left=6pt,
    right=6pt,
    top=6pt,
    bottom=6pt,
    width=\linewidth
]

\textbf{System Prompt:}
You are a helpful assistant.
\vspace{4pt}
\textbf{User Prompt:}
Please solve the following problem, strictly adhering to this format:

1.  **<think>...</think>**: Must contain the complete, step-by-step reasoning process. If information is missing, state why here and terminate your thinking process.

2.  **<answer>...</answer>**: Must ONLY contain the final answer enclosed in `\\boxed{}`. If information is insufficient, the answer must be 'insufficient information'.

\vspace{4pt}
\textbf{Input Format:}

\begin{verbatim}
Turn X:
State:
{problem input}
\end{verbatim}

\noindent
Always output: <think> [Your thoughts] </think> <answer> [your answer] </answer> with no extra text. Strictly follow this format.
\end{tcolorbox}
\caption{Prompt Example}
\label{tab:sys-prompt}
\end{table}
\subsection{Detailed Datasets Info Used in Analysis}
\label{app:data_report_sec5}

This appendix provides a detailed description of the datasets used in \S5, including their sources, construction procedures, and sizes.

\paragraph{Main experiments.}
We evaluate on two premise-missing benchmarks.
GSM8K-Insufficient is constructed from the GSM8K standard test set by removing essential premises, resulting in 865 premise-missing instances.
MetaMath-Insufficient is constructed from MetaMath in the same manner, resulting in 1130 premise-missing instances.

\paragraph{Analysis experiment 1.}
The data used in Analysis Experiment 1 is taken from the publicly available dataset released in Ref.~\cite{DBLP:journals/corr/abs-2504-06514}.

\paragraph{Analysis experiment 2.}
Analysis Experiment 2 uses standard information-complete test sets, including the GSM8K standard test set and the MATH500 test set.

\paragraph{Analysis experiment 3.}
Analysis Experiment 3 uses a mixed dataset containing both complete problems and premise-missing problems, with a total of 3814 instances.

\paragraph{Analysis experiment 4.}
Analysis Experiment 4 evaluates multi-domain QA transfer on premise-missing variants.
HotpotQA-Insufficient is constructed from HotpotQA, containing 1055 instances.
CQA-Insufficient is constructed from CommonsenseQA, containing 1170 instances.

\subsection{Llama-3B-Instruct training }
\begin{table}[htbp]
\centering
\footnotesize
\setlength{\tabcolsep}{8pt}
\renewcommand{\arraystretch}{1.05}
\begin{tabular}{@{}lcc cc@{}}
\toprule
& \multicolumn{2}{c}{\makecell[c]{\textbf{GSM8K-}\\\textbf{Insufficient}}}
& \multicolumn{2}{c}{\makecell[c]{\textbf{MetaMath-}\\\textbf{Insufficient}}} \\
\cmidrule(lr){2-3}\cmidrule(lr){4-5}
\makecell[l]{\textbf{Model}}
& \makecell{SR}
& \makecell{PD}
& \makecell{SR}
& \makecell{PD} \\
\midrule
Base Model         & 17.0 & 65.0 & 15.0 & 64.0 \\
w/ Prompt          & 43.8 & 80.8 & 42.9 & 73.6 \\
w/ Multi-turn RL   & 49.3 & 80.4 & 44.2 & 79.6 \\
\bottomrule
\end{tabular}
\caption{Performance on GSM8K-Insufficient and MetaMath-Insufficient}
\label{tab:baseline_comparison}
\end{table}

As shown in Table~\ref{tab:baseline_comparison} on Llama3-3B-Instruct~\cite{dubey2024llama}, prompting and multi-turn RL consistently
outperform the base model on both missing-premise benchmarks,
with GRIL achieving the best overall performance.
\section{The Use of Large Language Model(LLMs)}
To elevate the overall quality of this paper, a large language model was utilized to refine the manuscript—specifically, to enhance its clarity, streamline its conciseness, and ensure strict grammatical correctness.

\section{Case Study}
\subsection{Case Study 1}
\label{sec:case_before_training}

\begin{casebox}[]
\textbf{Case Study 1 (Before Training):} the model fails to proactively stop under insufficient information.

\vspace{6pt}
\textbf{System Prompt:} You are a helpful assistant.

\vspace{4pt}
\textbf{User Prompt:}
Please solve the following problem, strictly adhering to this format:
1. \texttt{<think>...</think>}: Must contain the complete, step-by-step reasoning process.
If information is missing, state why here and terminate your thinking process.
2. \texttt{<answer>...</answer>}: Must ONLY contain the final answer. If information is insufficient, the answer must be
\texttt{insufficient information}.
Always output: \texttt{<think> [Your thoughts] </think> <answer> [your answer] </answer>}
with no extra text. Strictly follow this format.

\vspace{6pt}
\textbf{Interaction Log:}\\
\textbf{system:}\\
You're a helpful assistant.

\textbf{user:}\\
Please solve the following problem, strictly adhering to this format:
1.  \textbf{<think>...</think>}: Must contain the complete, step-by-step reasoning process. If information is missing, state why here and terminate your thinking process.
2.  \textbf{<answer>...</answer>}: Must ONLY contain the final answer . If information is insufficient, the answer must be 'insufficient information'.

\textbf{Turn 1:}\\
\textbf{State:}\\
Find the sum of the remainders when the same integer is divided by 3 and by 5.
Always output: <think> [Your thoughts] </think> <answer> [your answer] </answer> with no extra text. Strictly follow this format.

\textbf{assistant:}\\
<think></think><answer>To solve the problem of finding the sum of the remainders when the same integer is divided by 3 and by 5, we need to understand the concept of modular arithmetic.

When an integer \( n \) is divided by 3, the remainder can be any of the following: 0, 1, or 2. Similarly, when \( n \) is divided by 5, the remainder can be any of the following: 0, 1, 2, 3, or 4.

The sum of the remainders when \( n \) is divided by 3 and by 5 will be the sum of all possible remainders for each divisor. Since there are 3 possible remainders for each divisor, we can calculate the total sum as follows:

Sum = (Remainder when divided by 3) + (Remainder when divided by 5)

There are 3 remainders when divided by 3 (0, 1, 2) and 5 remainders when divided by 5 (0, 1, 2, 3, 4). Therefore, the total sum of the remainders is:

Sum = 0 + 1 + 2 + 0 + 1 + 2 + 3 + 4

Sum = 15

So, the sum of the remainders when the same integer is divided by 3 and by 5 is 15.

 \(\boxed{15}\)</answer>

\textbf{user:}\\
Reward:\\
0.0

\textbf{Turn 2:}\\
\textbf{State:}\\
That is incorrect. Please try again.
Always output: <think> [Your thoughts] </think> <answer> [your answer] </answer> with no extra text. Strictly follow this format.

\textbf{assistant:}\\
<think>It appears there was a misunderstanding. Let me clarify the problem and provide the correct solution.</think><answer>When trying to find the sum of the remainders when the same integer is divided by 3 and by 5, we need to consider the possible remainders for each divisor. As previously stated, the remainders when divided by 3 are 0, 1, and 2, and the remainders when divided by 5 are 0, 1, 2, 3, and 4. The sum of these remainders is 15. Therefore, the correct answer is 15.</answer>

\textbf{user:}\\
Reward:\\
0.0

\textbf{Turn 3:}\\
\textbf{State:}\\
That is incorrect. Please try again.
Always output: <think> [Your thoughts] </think> <answer> [your answer] </answer> with no extra text. Strictly follow this format.

\textbf{assistant:}\\
<think>Let me rephrase the question and provide the correct solution.</think><answer>When a number is divided by 3, the possible remainders are 0, 1, and 2. When the same number is divided by 5, the possible remainders are 0, 1, 2, 3, and 4. The sum of these remainders is 15. Therefore, the correct answer is 15.</answer>

\textbf{user:}\\
Reward:\\
0.0

\textbf{Turn 4:}\\
\textbf{State:}\\
That is incorrect. Please try again.
Always output: <think> [Your thoughts] </think> <answer> [your answer] </answer> with no extra text. Strictly follow this format.

\textbf{assistant:}\\
<think>Let me restate the problem clearly and provide the correct solution.</think><answer>The sum of the remainders when a number is divided by 3 and 5 is 15. This is because the possible remainders when divided by 3 are 0, 1, and 2, and when divided by 5, they are 0, 1, 2, 3, and 4. Adding these together gives 15.</answer>
\vspace{6pt}
\textbf{Observation.}
The problem is underdetermined since the integer is unspecified. However, the model repeatedly attempts to solve the problem instead of detecting the missing premise and stopping or requesting clarification.

\end{casebox}

\begin{casebox}[]

\textbf{Case Study 1 (After Training):}
The model proactively stops, requests missing premises, and solves after clarification.

\vspace{6pt}
\textbf{System Prompt:}

You're a helpful assistant.

\vspace{4pt}
\textbf{User Prompt:}

Please solve the following problem, strictly adhering to this format:
\begin{itemize}
  \item \texttt{<think>...</think>}: Must contain the complete, step-by-step reasoning process. If information is missing, state why here and terminate your thinking process.
  \item \texttt{<answer>...</answer>}: Must ONLY contain the final answer . If information is insufficient, the answer must be \texttt{insufficient information}.
\end{itemize}

Always output:
\texttt{<think> [Your thoughts] </think> <answer> [your answer] </answer>}
with no extra text.

\vspace{6pt}
\textbf{Interaction Log:}

\textbf{Turn 1:}\\
\textbf{State:}\\
Find the sum of the remainders when the same integer is divided by 3 and by 5.
Always output: <think> [Your thoughts] </think> <answer> [your answer] </answer> with no extra text. Strictly follow this format.

\textbf{assistant:}\\
<think>Let's break down the problem step by step. We need to find the sum of the remainders when the same integer is divided by 3 and by 5. However, without knowing the specific value of the integer, it is impossible to determine the remainders. Therefore, the problem does not provide enough information to determine the remainders.</think><answer>Insufficient information</answer>

\textbf{user:}\\
Reward:\\
1.0

\textbf{Turn 2:}\\
\textbf{State:}\\
Successfully detected a missing premise. Here is the missed information: When an integer is divided by 15, the remainder is 7.. Please solve the problem now.You should give detailed reasonig steps
Always output: <think> [Your thoughts] </think> <answer> [your answer] </answer> with no extra text. Strictly follow this format.

\textbf{assistant:}\\
<think>Let's break down the problem step by step. We know that the remainder when an integer is divided by 15 is 7. Therefore, the integer can be written in the form \( N = 15k + 7 \) for some integer \( k \). We need to find the remainders when \( N \) is divided by 3 and by 5. The remainder when \( N \) is divided by 3 is \( 7 \mod 3 = 1 \). The remainder when \( N \) is divided by 5 is \( 7 \mod 5 = 2 \). Therefore, the sum of the remainders is \( 1 + 2 = 3 \). Therefore, the sum of the remainders is 3.</think><answer>3</answer>

\vspace{6pt}
\textbf{Observation.}
After training, the model correctly identifies the underdetermined input and stops with an
\texttt{insufficient information} response. Once the missing premise is provided, it
successfully integrates the new condition and solves the problem in the next turn.

\end{casebox}

\subsection{Case Study 2}
\label{sec:case_before_training_2}

\begin{casebox}
\textbf{Case Study 2 (Before Training):}
The model repeatedly attempts to solve an ill-defined problem without identifying the missing premise.

\vspace{6pt}
\textbf{System Prompt:}

You're a helpful assistant.

\vspace{4pt}
\textbf{User Prompt:}

Please solve the following problem, strictly adhering to this format:
\begin{itemize}
  \item \texttt{<think>...</think>}: Must contain the complete, step-by-step reasoning process. If information is missing, state why here and terminate your thinking process.
  \item \texttt{<answer>...</answer>}: Must ONLY contain the final answer. If information is insufficient, the answer must be \texttt{insufficient information}.
\end{itemize}

Always output:
\texttt{<think> [Your thoughts] </think> <answer> [your answer] </answer>}
with no extra text.

\vspace{6pt}
\textbf{Interaction Log:}
\textbf{Turn 1:}\\
\textbf{State:}\\
Barbara got a great deal on a new chest of drawers, but she has to take a lot of paper
out of the drawers to be able to use it. If a bunch holds 4 sheets of paper, a bundle
holds 2 sheets of paper, and a heap holds 20 sheets of paper, how many sheets of paper
did Barbara remove from the chest of drawers? If we know the answer to the above
question is 114, what is the value of unknown variable x?

\textbf{assistant:}\\
<think>To solve this problem, we need to determine the value of the unknown variable x,
which represents the number of drawers in the chest of drawers.

From the given information:
- A bunch holds 4 sheets of paper.
- A bundle holds 2 sheets of paper.
- A heap holds 20 sheets of paper.

We also know that Barbara removed a total of 114 sheets of paper.

We can set up an equation:
4x + 2x + 20x = 114

Now, we can solve this equation.</think>
<answer>x = 5</answer>

Reward:\\
0.0

\textbf{Turn 2:}\\
\textbf{State:}\\
That is incorrect. Please try again.

\textbf{assistant:}\\
<think>I apologize for the mistake. Let's re-examine the problem.

We can set up the equation:
4x + 2x + 20x = 114</think>
<answer>5</answer>

Reward:\\
0.0

\textbf{Turn 3:}\\
\textbf{State:}\\
That is incorrect. Please try again.

\textbf{assistant:}\\
<think>I apologize for the mistake. Let's re-examine the problem.

We can set up the equation:
4x + 2x + 20x = 114</think>
<answer>5</answer>

Reward:\\
0.0

\textbf{Turn 4:}\\
\textbf{State:}\\
That is incorrect. Please try again.

\textbf{assistant:}\\
<think>I apologize for the mistake. Let's re-examine the problem.

We can set up the equation:
4x + 2x + 20x = 114</think>
<answer>5</answer>

\vspace{6pt}
\textbf{Observation.}
Despite repeated negative feedback, the model fails to identify the missing or ill-defined
premise regarding the variable $x$ and persistently attempts to solve the problem using
an incorrect formulation, illustrating overconfident reasoning and the absence of a
proactive stopping strategy.

\end{casebox}

\label{sec:case_after_training_2}

\begin{casebox}

\textbf{Case Study 2 (After Training):}
The model proactively detects missing premises and successfully solves the task after clarification.

\vspace{6pt}
\textbf{System Prompt:}

You're a helpful assistant.

\vspace{4pt}
\textbf{User Prompt:}

Please solve the following problem, strictly adhering to this format:
1. \texttt{<think>...</think>}: Must contain the complete, step-by-step reasoning process. If information is missing, state why here and terminate your thinking process.
2. \texttt{<answer>...</answer>}: Must ONLY contain the final answer. If information is insufficient, the answer must be \texttt{insufficient information}.

Always output: \texttt{<think> [Your thoughts] </think> <answer> [your answer] </answer>}
with no extra text. Strictly follow this format.

\vspace{6pt}
\textbf{Interaction Log:}
\textbf{system:}\\
You're a helpful assistant.

\textbf{user:}\\
Please solve the following problem, strictly adhering to this format:
1.  \textbf{<think>...</think>}: Must contain the complete, step-by-step reasoning process. If information is missing, state why here and terminate your thinking process.
2.  \textbf{<answer>...</answer>}: Must ONLY contain the final answer . If information is insufficient, the answer must be 'insufficient information'.

\textbf{Turn 1:}\\
\textbf{State:}\\
Barbara got a great deal on a new chest of drawers, but she has to take a lot of paper out
of the drawers to be able to use it. If a bunch holds 4 sheets of paper, a bundle holds 2
sheets of paper, and a heap holds 20 sheets of paper, how many sheets of paper did Barbara
remove from the chest of drawers? If we know the answer to the above question is 114, what
is the value of unknown variable x?
Always output: <think> [Your thoughts] </think> <answer> [your answer] </answer> with no
extra text. Strictly follow this format.

\textbf{assistant:}\\
<think>Let's break down the total number of sheets of paper in the chest of drawers. We know
the total number of sheets removed, which is 114. We need to determine the total number of
sheets of paper in the chest of drawers. Since each bunch holds 4 sheets, each bundle holds
2 sheets, and each heap holds 20 sheets, we can express the total number of sheets as the sum
of the number of bunches, bundles, and heaps. Since the problem does not provide the number of
bunches, bundles, and heaps, we cannot determine the exact value of the unknown variable x.
Therefore, the problem does not provide enough information to determine the value of the unknown
variable x.</think><answer>Insufficient information</answer>

\textbf{user:}\\
Reward:\\
1.0

\textbf{Turn 2:}\\
\textbf{State:}\\
Successfully detected a missing premise. Here is the missed information: She found x bundles
of colored paper, 2 bunches of white paper, and 5 heaps of scrap paper.. Please solve the
problem now.You should give detailed reasonig steps
Always output: <think> [Your thoughts] </think> <answer> [your answer] </answer> with no
extra text. Strictly follow this format.

\textbf{assistant:}\\
<think>Let's break down the total number of sheets of paper in the chest of drawers. We know
the total number of sheets removed, which is 114. We need to determine the total number of
sheets of paper in the chest of drawers. The total number of bundles is x, each bundle holds
2 sheets, so the total number of sheets in bundles is 2x. The total number of bunches is 2,
each bunch holds 4 sheets, so the total number of sheets in bunches is 2 * 4 = 8. The total
number of heaps is 5, each heap holds 20 sheets, so the total number of sheets in heaps is
5 * 20 = 100. The total number of sheets is 2x + 8 + 100 = 2x + 108. The total number of
sheets removed is 114. Therefore, 2x + 108 = 114. Solving for x, we get 2x = 114 - 108 = 6.
Therefore, x = 6 / 2 = 3.</think><answer>3</answer>

\vspace{6pt}
\textbf{Observation.}
After training, the model correctly stops when the initial question is under-specified and
requests missing premises. Once the additional information is provided, it integrates the
new constraints and solves for the unknown variable in the subsequent turn.

\end{casebox}
\subsection{Case of GapRatio}
\begin{tcolorbox}[
    colback=white,
    colframe=black,
    boxrule=1pt,
    left=6pt,
    right=6pt,
    top=6pt,
    bottom=6pt,
    width=\linewidth,
    breakable
]
Okay, so Lani baked 55 cookies, and the question is asking how many cookies were in each jar. Hmm, let me think about this. First off, I need to figure out what information is given and what's missing. We know that Lani baked 55 cookies, 
\tcbox[on line, colframe=red!60!black, colback=white, boxrule=1pt, size=fbox, arc=2pt]{but we don't know anything else about the jars.} 
Maybe she put them in jars, but how many jars? Was each jar filled equally? Is there a specific number of jars mentioned?I don't think so. The problem doesn't specify how many jars there are or if the cookies were divided equally among them. Without that information, it's hard to determine how many cookies are in each jar. If there were, say, 5 jars, then each jar would have 11 cookies. But since we don't know the number of jars, we can't make that assumption.Maybe the question expects me to assume something, like maybe it's divided into jars with the same number of cookies, but even then, without knowing the total number of jars, I can't calculate the exact number per jar. It's like trying to split something without knowing how many parts it's being split into.Wait, maybe the question is trying to test if I recognize that information is missing? So, maybe the answer is that there's insufficient information to determine the number of cookies per jar. Because without knowing how many jars there are, we can't figure out how the cookies are distributed.Yeah, that makes sense. So, the problem is underspecified because it doesn't provide enough details about the number of jars or any other factors that could help determine the distribution. Without that, it's impossible to know how many cookies are in each jar.So, I think the answer is that there's not enough information provided to determine the number of cookies in each jar.</think>The problem states that Lani baked 55 cookies but does not specify the number of jars or any other details about how the cookies were distributed. Without this information, we cannot determine how many cookies were in each jar.\\

\textbf{Observation}.
By the third sentence, the model has already begun to suspect that we do not know any information about the jars. However, after recognizing this issue, the model does not choose to stop; instead, it repeatedly engages in doubt and speculation, for example by assuming that there are five jars. This example illustrates the evidence-free reasoning exhibited by an untrained model when confronted with missing-premise problems, which in turn leads to a high GapRatio.
\end{tcolorbox}

\end{document}